\documentclass[journal]{IEEEtran}
\usepackage{cite}
\usepackage[pdftex]{graphicx}
\DeclareGraphicsExtensions{.pdf,.jpg,.png}
\usepackage[pdfborder=false,colorlinks,bookmarksnumbered,bookmarksopen,linkcolor=black,citecolor=black,urlcolor=black]{hyperref}
\usepackage[tight,footnotesize]{subfigure}
\usepackage{rotating}
\usepackage{graphicx}
\usepackage{amsmath,amssymb} 
\usepackage{epsfig}
\usepackage{array}
\usepackage{multirow}
\usepackage{colortbl}
\usepackage{amsfonts}

\usepackage{pifont}

\definecolor{gray1}{rgb}{.8,.8,.8}
\usepackage{xspace}
\usepackage{footnote}
\usepackage{etoolbox}
\usepackage{overpic}
\usepackage{multirow}
\usepackage{multicol}

\def\ie{\emph{i.e.}}
\def\eg{\emph{e.g.}}
\def\etal{{\em et al.}}

\newcommand{\first}[1]{{\textcolor{red}{#1}}}
\newcommand{\second}[1]{{\textcolor{green}{#1}}}
\newcommand{\third}[1]{{\textcolor{blue}{#1}}}
\newcommand{\figref}[1]{Fig. \ref{#1}}
\newcommand{\tabref}[1]{Fig. \ref{#1}}
\newcommand{\equref}[1]{(\ref{#1})}
\newcommand{\secref}[1]{Sec. \ref{#1}}

\renewcommand{\arraystretch}{1.1}
\renewcommand{\tabcolsep}{.5mm}

\newcommand{\Cols}[1]{\multicolumn{3}{c|}{#1}}
\newcommand{\tabTitle}{}

\newcommand{\AddImg}[1]{}
\newcommand{\AddImgs}[2]{}
\newcommand{\AddImgsU}[2]{}
\newcommand{\myPara}[1]{\vspace{.05in}\noindent\textbf{#1.}}

\graphicspath{{./figs/}}

\usepackage{color}
\ifCLASSINFOpdf
\else
\fi

\begin{document}
\title{Salient Object Detection: A Benchmark}

\author{Ali~Borji,~Ming--Ming~Cheng,~Huaizu Jiang~and~Jia Li%
\thanks{Manuscript received January 5, 2015; 
revised July 13, 2015 and September 19, 2015; 
accepted October 4, 2015. Date of publication October 7, 2015; 
date of current version October 23, 2015. 
The associate editor coordinating the review of this manuscript 
and approving it for publication was Prof. Christine Guillemot. 
(Ali Borji and Ming-Ming Cheng equally contributed to this work.)}
\thanks{A. Borji is with the Computer Science Department,
University of Wisconsin, Milwaukee, WI 53211. E-mail: borji@uwm.edu}
\thanks{M.M Cheng (corresponding author) is with CCCE \& CS, 
Nankai University, Jinnan, 
Tianjin, P.R.China, 300353. E-mail: cmm@nankai.edu.cn}
\thanks{H. Jiang is with the Institute of Artificial Intelligence
and Robotics, Xi'an Jiaotong University, China.
E-mail: hzjiang@mail.xjtu.edu.cn}
\thanks{J. Li is with State Key Laboratory of Virtual Reality Technology
and Systems, School of Computer Science and Engineering,
Beihang University. He is also with the International Research Institute
for Multidisciplinary Science (IRIMS) at Beihang University,
Beijing, China. E-mail: jiali@buaa.edu.cn}
\thanks{An earlier version of this work has been published in ECCV
2012~\cite{borji2012salient}.}
\thanks{Please contact Ming-Ming Cheng as corresponding author for 
benchmark updates at \href{http://mmcheng.net/salobjbenchmark}{http://mmcheng.net/salobjbenchmark}.}
\thanks{Color versions of one or more of the figures in this paper are available online at http://ieeexplore.ieee.org.}
\thanks{Digital Object Identifier 10.1109/TIP.2015.2487833}}

\markboth{IEEE TRANSACTIONS ON IMAGE PROCESSING, VOL. 24, NO. 12, DECEMBER 2015}{Shell \MakeLowercase{\textit{et al.}}: Bare Demo of IEEEtran.cls for Journals}

\maketitle

\begin{abstract}
We extensively compare, qualitatively and quantitatively, 41 state-of-the-art models (29 salient object detection, 10 fixation prediction, 1 objectness, and 1 baseline) over 7 challenging datasets for the purpose of benchmarking salient object detection and segmentation methods.
From the results obtained so far, our evaluation shows a consistent rapid progress over the last few years in terms of both accuracy and running time. The top contenders in this benchmark significantly outperform the models identified as the best in the previous benchmark conducted three years ago. We find that the models designed specifically for salient object detection generally work better than models in closely related areas, which in turn provides a precise definition and suggests an appropriate treatment of this problem that distinguishes it from other problems. In particular, we analyze the influences of center bias and scene complexity in model performance, which, along with the hard cases for state-of-the-art models, provide useful hints towards constructing more challenging large scale datasets and better saliency models. Finally, we propose probable solutions for tackling several open problems such as evaluation scores and dataset bias, which also suggest future research directions in the rapidly-growing field of salient object detection.
\end{abstract}


\IEEEpeerreviewmaketitle

\section{Introduction}

\IEEEPARstart{V}ISUAL \emph{attention}, the astonishing capability of human visual system to selectively process only the \emph{salient} visual stimuli in details, has been investigated by multiple disciplines such as cognitive psychology, neuroscience, and computer vision
\cite{borji2013state,borji2013quantitative,hayhoe2005eye,itti2001computational}.
Following cognitive theories (\eg,~\textit{feature integration theory (FIT)}~\cite{treisman1980feature}, \textit{guided search model} \cite{wolfe1989guided,wolfe2005guidance}) and early attention models (\eg, Koch and Ullman~\cite{koch1987shifts} and Itti~\etal~\cite{itti1998model}), hundreds of computational saliency models have been proposed to detect salient visual subsets from images and videos.

Despite the psychological and neurobiological definitions, the concept of visual saliency is becoming vague in the field of computer vision. Some visual saliency models (\eg, \cite{itti1998model,parkhurst2002modeling,li2010probabilistic,borji2012exploiting,borji2012boosting,borji2013quantitative,koehler2014saliency,li2014visual}) aimed to \emph{predict human fixations} as a way to test their accuracy in saliency detection, while other models~\cite{LiuSZTS07Learn,achanta2009frequency,tian2014learning}, which were often driven by computer vision applications such as content-aware image resizing and photo visualization~\cite{wang2006picture}, attempted to \emph{identify salient regions/objects} and used explicit saliency judgments for evaluation~\cite{borji2013stands}. Although both types of saliency models are expected to be applicable interchangeably, their generated saliency maps actually demonstrate remarkably different characteristics due to the distinct purposes in saliency detection. For example, fixation prediction models usually pop-out sparse blob-like salient regions, while salient object detection models often generate smooth connected areas. On the one hand, detecting large salient areas often causes severe false positives for fixation prediction.
On the other hand, popping-out only sparse salient regions causes massive misses in detecting salient regions and objects.


To separate these two types of saliency models, in this study we provide a precise definition and suggest an appropriate treatment of salient object detection. Generally, a salient object detection model should, \textit{first} detect the salient attention-grabbing objects in a scene, and \textit{second}, segment the entire objects. Usually, the output of the model is a saliency map where the intensity of each pixel represents its probability of belonging to salient objects. From this definition, we can see that this problem in its essence is a figure/ground segmentation problem, and the goal is to only segment the salient foreground object
from the background. Note that it slightly differs from the traditional image segmentation problem that aims to partition an image into perceptually coherent regions.





The value of salient object detection models lies in their applications in many areas such as computer vision, graphics, and robotics.
For instance, these models have been successfully applied in many applications such as object detection and recognition~\cite{rutishauser2004bottom,kanan2010robust,moosmann2006learning,borji2011cost,borji2011scene,shen2013moving,borji2014salient,ren2013region,guo2014fast}, image and video compression~\cite{guo2010novel,itti2004automatic},
video summarization~\cite{ma2005generic,lee2012discovering,ji2012video},
photo collage/media re-targeting/cropping/thumb-nailing~\cite{goferman2010puzzle,wang2006picture,huang2011arcimboldo},
image quality assessment~\cite{ninassi2007does,liu2009studying,li2013color,zhang2015application},
image segmentation~\cite{donoser2009saliency,li2011saliency,qin2013integration,johnson2010attention},
content-based image retrieval and image collection browsing
\cite{chen2009sketch2photo,feng2010attention,sun2013image,li2013partial},
image editing and manipulating~
\cite{chia2011semantic,liu2012web,margolin2013saliency,goldberg2012data},
visual tracking
\cite{stalder2013dynamic,li2013visual,garcia2012adaptive,borji2012adaptive,klein2010adaptive,frintrop2009most,zhang2010visual},
object discovery~\cite{karpathyobject,frintropcognitive},
and human-robot interaction~\cite{meger2008curious,sugano2010calibration,borji2014defending}. 

The field of salient object detection develops very fast. Many new models and benchmark datasets have been proposed since our earlier benchmark conducted three years ago~\cite{borji2012salient}. Yet, it is unclear how the new algorithms fare against previous models and new datasets. Are there any \emph{real improvements} in this field or we are just fitting models to datasets? It is also interesting to test the performance of old high-performing models on the new benchmark datasets. A recent exhaustive review of salient object detection models can be found in~\cite{borji2014salient}.



In this study, we compare and analyze models from three categories: 1) salient object detection, 2) fixation prediction, and
3) object proposal generation\footnote{Object proposal generation is a recently emerging trend which attempts to detect image regions that may contain objects from any object category (a.k.a, category independent object proposals).}.
The reason to include the latter two types of models is to conduct across-category comparison and to study whether models specifically
designed for salient object detection show actual advantage over models for fixation prediction and object proposal generation.
This is particularly important since these models have different objectives and generate visually distinctive maps.
We also include a baseline model to study the effect of center bias in model comparison. In summary, we hope that such a benchmark not only allows researchers to compare their models with other algorithms but also helps identify the chief factors affecting the performance of salient object detection models.
%

\section{Salient Object Detection Benchmark}
\label{sec:benchmark}
In this benchmarking, we focus on evaluating models whose input is a single image. This is due to the fact that salient object detection on a single input image is the main research direction, while the comprehensive evaluation of models working on multiple input images (\eg, co-salient object detection and spatio-temporal saliency) lacks public benchmarks.

\subsection{Compared Models}
\begin{figure}[t]
  \centering
  \vspace{-.07in}
  \subfigure[\textbf{MSRA10K}]{\includegraphics[width=.32\linewidth]{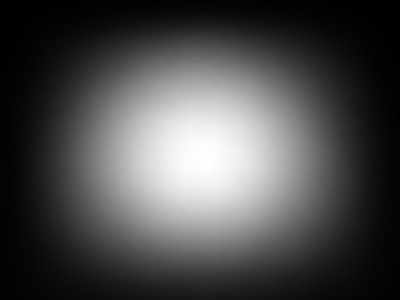}}
  \subfigure[\textbf{PASCAL-S}]{\includegraphics[width=.32\linewidth]{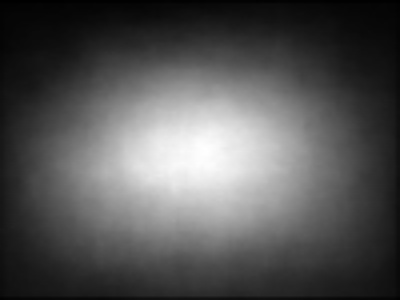}}
  \subfigure[\textbf{THUR15K}]{\includegraphics[width=.32\linewidth]{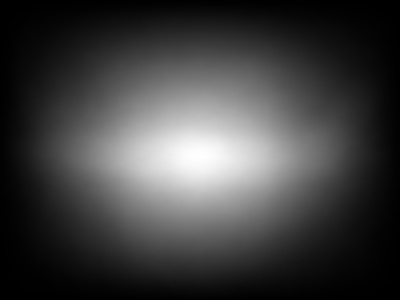}}\vspace{-.08in}
  \subfigure[\textbf{DUT-OMRON}]{\includegraphics[width=.32\linewidth]{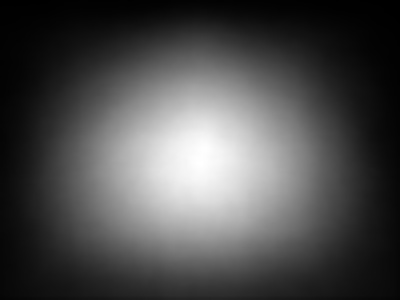}}
  \subfigure[\textbf{JuddDB}]{\includegraphics[width=.32\linewidth]{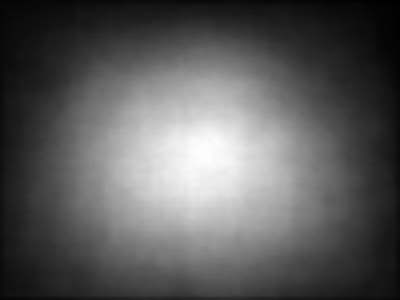}}
  \subfigure[\textbf{SED2}]{\includegraphics[width=.32\linewidth]{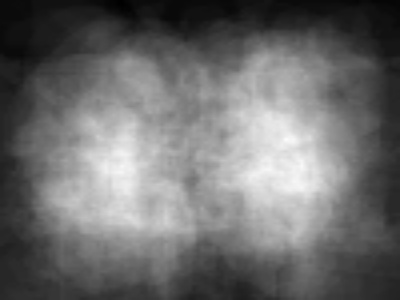}}
  \caption{Average annotation maps of six datasets used in benchmarking.
  }\label{fig:AverageMap}
\end{figure}

In this study, we run 41 models in total (29 salient object detection models, 10 fixation prediction models, 1 objectness proposal model, and 1 baseline) whose codes or executables were accessible (see Fig.~\ref{tab:salientObjModelsIntrin} for a complete list). The baseline model, denoted as ``Average Annotation Map (AAM),'' is simply the average of ground-truth annotations of all images on each dataset. Note that AAM often has a larger activation at the image center (see \figref{fig:AverageMap}), and we can thus study the effect of center bias in model comparison.

\begin{figure}[t]
  \centering
  \footnotesize
  \renewcommand{\arraystretch}{1.0}
  \renewcommand{\tabcolsep}{1.8mm}
  \begin{tabular}{|c||lcc|c|c|c|c|}
  \hline
  \# & \textbf{Model} & \textbf{Pub} & \textbf{Year} &  \textbf{Code} & \textbf{Time(s)} & \textbf{Cat.}   \\
  \hline \hline
  1 & \textbf{AC}~\cite{achanta2008salient} & ICVS & 2008 & C & .129 & \textbf{\multirow{38}{*}{\begin{rotate}{90}Salient Object Detection\end{rotate}}}\\
  2 & \textbf{FT}~\cite{achanta2009frequency} & CVPR & 2009 & C & .072 & \\
  3 & \textbf{CA}~\cite{goferman2012context} & CVPR & 2010 & M + C & 40.9 &\\
  4 & \textbf{MSS}~\cite{Achanta10saliency} & ICIP & 2010 & C & .076 & \\
  5 & \textbf{SEG}~\cite{rahtu2010segmenting} & ECCV & 2010 & M + C & 10.9 & \\
  6 & \textbf{RC}~\cite{ChengPAMI} & CVPR & 2011 &  C & .136 & \\
  7 & \textbf{HC}~\cite{ChengPAMI} & CVPR & 2011 & C & .017 & \\
  8 & \textbf{SWD}~\cite{duan2011visual} & CVPR & 2011 & M + C & .190 & \\
  9 & \textbf{SVO}~\cite{chang2011fusing} & ICCV & 2011 &  M + C & 56.5 & \\
  10 & \textbf{CB}~\cite{jiang2011automatic} & BMVC & 2011 &  M + C & 2.24 & \\
  11 & \textbf{FES}~\cite{Tavakoli11fast} & Img.Anal. & 2011 & M + C & .096 &\\
  12 & \textbf{SF}~\cite{perazzi2012saliency} & CVPR & 2012 &  C & .202 &\\
  13 & \textbf{LMLC}~\cite{XieEtAlTIP2013} & TIP & 2013 & M + C & 140. &\\
  14 & \textbf{HS}~\cite{yan2013hierarchical} & CVPR & 2013 & EXE & .528 &\\
  15 & \textbf{GMR}~\cite{YangZLRY13Manifold} & CVPR & 2013 & M & .149 &\\
  16 & \textbf{DRFI}~\cite{JiangWYWZL13} & CVPR & 2013 & C & .697 &\\
  17 & \textbf{PCA}~\cite{margolinmakes} & CVPR & 2013 & M + C & 4.34 & \\
  18 & \textbf{LBI}~\cite{siva2013looking} & CVPR & 2013 & M + C & 251. & \\
  19 & \textbf{GC}~\cite{ChengWLZVC13Efficient} & ICCV & 2013 & C & .037 & \\
  20 & \textbf{CHM}~\cite{LiLSDH13Contextual} & ICCV & 2013 & M + C & 15.4 & \\
  21 & \textbf{DSR}~\cite{li2013saliency} & ICCV & 2013 & M + C & 10.2 & \\
  22 & \textbf{MC}~\cite{Jiang2013Saliency} & ICCV & 2013 & M + C & .195 &\\
  23 & \textbf{UFO}~\cite{JiangLYP13UFO} & ICCV & 2013 & M + C & 20.3 &\\
  24 & \textbf{MNP}~\cite{margolin2013saliency} & Vis.Comp. & 2013 & M + C & 21.0 &\\
  25 & \textbf{GR}~\cite{yang2013graph} & SPL & 2013 & M + C & 1.35 &\\
  26 & \textbf{RBD}~\cite{zhu2014saliency} & CVPR & 2014 & M & .269 &\\
  27 & \textbf{HDCT}~\cite{kim2014salient} & CVPR & 2014 & M & 4.12 &\\
  28 & \textbf{ST}~\cite{liu2013saliencyTIP} & TIP & 2014 & M+C & 79.1 &\\
  29 & \textbf{QCUT}~\cite{aytekin2014automatic} & ICPR & 2014 & M + C & 1.82 &\\

  \hline \hline
   1 & \textbf{IT}~\cite{itti1998model}    & PAMI & 1998  & M & .302 & \textbf{\multirow{17}{*}{\begin{rotate}{90}Fixation Prediction\end{rotate}}} \\
   2 & \textbf{AIM}~\cite{bruce2005saliency} & JOV &2006  & M & 8.66 &\\
   3 & \textbf{GB}~\cite{harel2007graph}  & NIPS&2007  & M + C  & .735 & \\
   4 & \textbf{SR}~\cite{hou2007saliency}       & CVPR&2007  & M & .040 &\\
   5 & \textbf{SUN}~\cite{zhang2008sun}    & JOV &2008  &M & 3.56 &\\
   6 & \textbf{SeR}~\cite{seo2009static}      & JOV & 2009 & M & 1.31 &\\
   7 & \textbf{SIM}~\cite{murray2011saliency}&CVPR&2011& M & 1.11 &\\
   8 & \textbf{SS}~\cite{hou2012image}      & PAMI &2012 & M & .053 &\\
   9 & \textbf{COV}~\cite{erdem2013visual}     & JOV  &2013 & M & 25.4 &\\
   10& \textbf{BMS}~\cite{zhang2013boolean}& ICCV &2013 & M + C & .575 &\\
  \hline \hline
   1 & \textbf{OBJ}~\cite{alexe2010object}  & CVPR & 2010 &  M+C & 3.01 & - \\
\hline \hline
   1 & \textbf{AAM}  & - & - &  - & - & - \\
  \hline
  \end{tabular}
  \caption{Compared salient object detection,
    fixation prediction, object proposal generation, and
    baseline models sorted by their publication year \{M= Matlab, C= C/C++, EXE = executable\}.
    The average running time is tested on MSRA10K dataset (typical
    image resolution $400\times 300$) using a desktop machine with Xeon
    E5645 2.4 GHz CPU and 8GB RAM.
    We evaluate those models whose codes or executables are available.
  }\label{tab:salientObjModelsIntrin}
\end{figure}

\subsection{Datasets}

Since there exist many datasets that differ in number of images, number of objects per image, image resolution and annotation form (bounding box or accurate region mask), it is likely that models may rank differently across datasets.
Hence, to come up with a fair comparison, it is necessary to run models over multiple datasets so as to draw objective conclusions.
A good model should perform well over almost all datasets. Toward this end, seven datasets\footnote{To save space, we show some plots over the \textbf{ECSSD} dataset on our online benchmark website.} were chosen for model comparison, including:
1) \textbf{MSRA10K}~\cite{THUR15000db}, 2) \textbf{THUR15K}~\cite{THUR15000db}, 3) \textbf{ECSSD}~\cite{yan2013hierarchical}, 4) \textbf{JuddDB}~\cite{borjiTIP2014}, 5) \textbf{DUT-OMRON}~\cite{YangZLRY13Manifold} and 6) \textbf{SED2}~\cite{alpert2007image,borji2012salient}, and
7) \textbf{PASCAL-S}~\cite{liXiaodiCVPR2014}. These datasets were selected based on the following four criteria: 1) being widely-used, 2) containing a large number of images,
3) having different biases (\eg, number of salient objects, image clutter, center-bias),
and 4) potential to be used as benchmarks in the future research.

\textbf{MSRA10K} is a descendant of the MSRA dataset~\cite{LiuSZTS07Learn}. It  contains 10,000 annotated images that covers all the 1,000 images in the popular ASD dataset~\cite{achanta2009frequency}.
\textbf{THUR15K} and \textbf{DUT-OMRON} are used to compare models on a large scale.
\textbf{ECSSD} contains a large number of semantically meaningful but structurally complex natural images.
The reason to include \textbf{JuddDB} and \textbf{PASCAL-S} datasets was to assess performance of models over scenes with multiple objects with high background clutter. Finally, we also evaluate models over \textbf{SED2} to check whether salient object detection algorithms can perform well on images containing more than one salient object (\ie, two in \textbf{SED2}).
\figref{fig:AverageMap} shows the AAM model output of six benchmark datasets to illustrate their different center biases. See~\figref{fig:SampleImgs} for representative images and annotations from each dataset.

\begin{figure*}[t]
    \centering	
    \includegraphics[width=\linewidth]{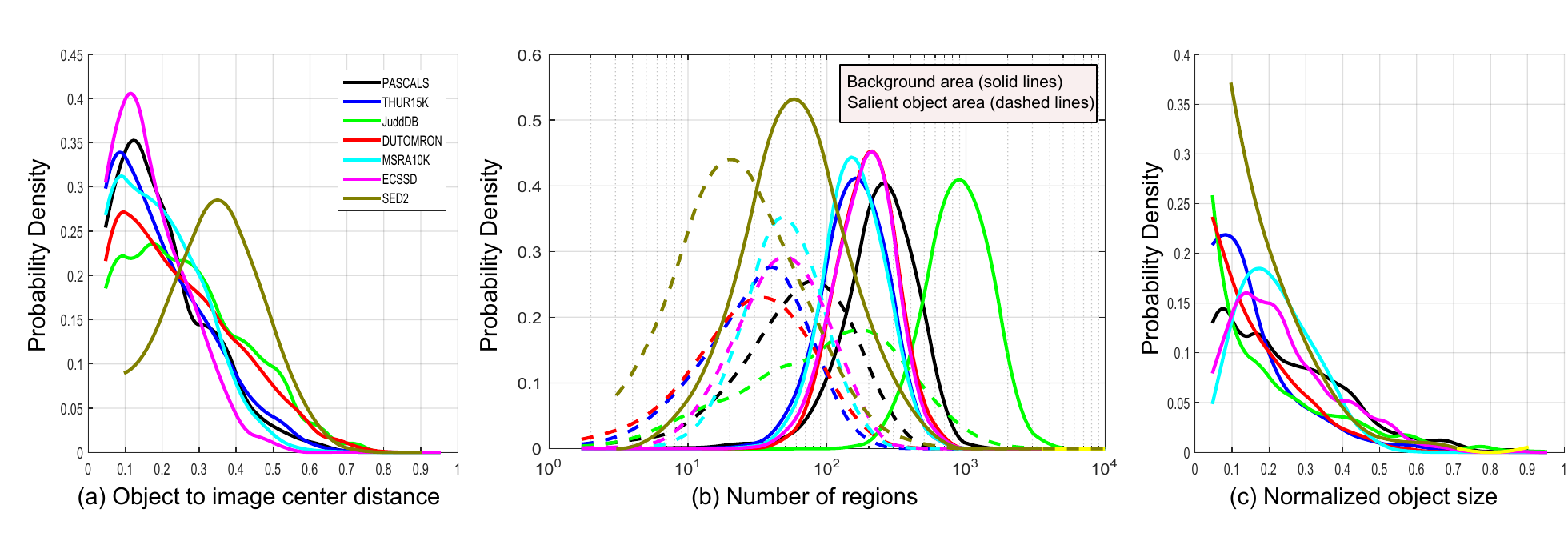} \\
    \caption{Statistics of the benchmark datasets.
        a) distribution of normalized object distance from image center,
        b) distribution of number of super-pixels on salient objects and image background, and
        c) distribution of normalized object size. See text for precise definitions.
    }\label{fig:dbStats}
\end{figure*}

We illustrate in \figref{fig:dbStats} the statistics of the seven chosen datasets.
In \figref{fig:dbStats}(a), we show the normalized distances from the centroid
of salient objects to the corresponding image centers.
We can see that salient objects in \textbf{ECCSD} have the shortest
distance to image centers, while salient objects in \textbf{SED2} have the longest distances.
This is reasonable since images in \textbf{SED2} usually have two objects
aligned around opposite image borders.
Moreover, we can see that the spatial distribution of salient objects in \textbf{JuddDB}
has a larger variety than other datasets,
indicating that this dataset has smaller positional bias
(\ie, center-bias of salient objects and border-bias of background regions).

In \figref{fig:dbStats}(b), we aim to show the complexity of images in seven
benchmark datasets.
Toward this end, we apply the segmentation algorithm by Felzenszwalb
\etal~\cite{felzenszwalb2004efficient} to see how many super-pixels
(\ie, homogeneous regions) can be obtained on average from salient objects
and background regions of each image, respectively.
In this manner, we can use this measure to reflect how challenging a benchmark dataset is since massive super-pixels often indicate complex foreground objects and cluttered background.
From \figref{fig:dbStats}(b), we can see that \textbf{JuddDB} (followed by \textbf{PASCAL-S}) is the most challenging
benchmark since it has an average number of 493 super-pixels from the background of each image.
On the contrary, \textbf{SED2} contains fewer number of super-pixels in
foreground and background regions,
indicating that images in this benchmark often contain uniform regions and are relatively easier to process.

\newcommand{\addSampleImgs}[5]{{\includegraphics[height=#2\linewidth]{SampleImgs/#3/#4.#1}
    \includegraphics[height=#2\linewidth]{SampleImgs/#3/#5.#1}}}
\newcommand{\addSampleImgsAndGt}[6]{\begin{minipage}[b]{#6\linewidth}\addSampleImgs{jpg}{#1}{#2}{#3}{#4} \\
    \addSampleImgs{png}{#1}{#2}{#3}{#4} \\ \makebox[\textwidth][c]{#5} \end{minipage}}
\begin{figure}[t]
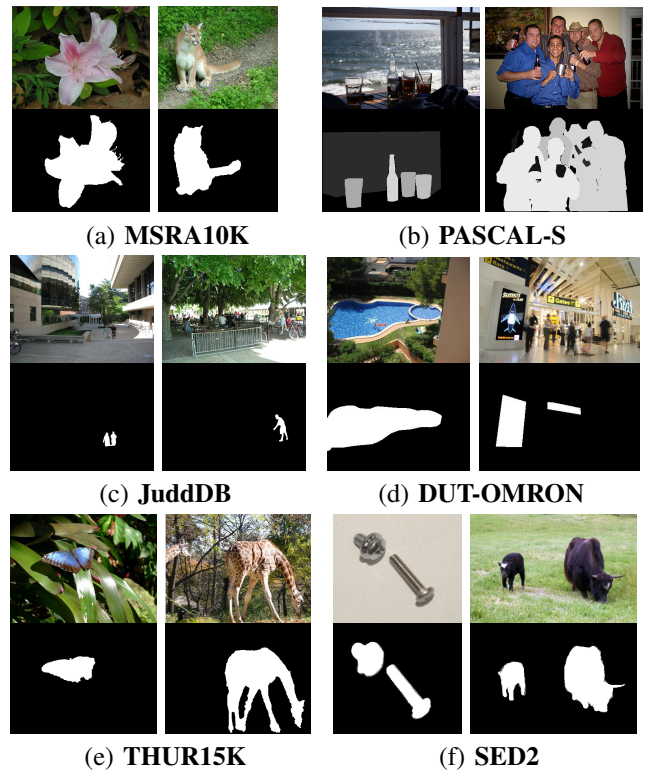

  \centering
  \addSampleImgsAndGt{.33}{MSRA}{187033}{85416}{(a) \textbf{MSRA10K}}{.49} \hspace{-.1in}
  \addSampleImgsAndGt{.32}{PASCAL}{341}{397}{(b) \textbf{PASCAL-S}}{.504}\\ \vspace{.04in}
  \addSampleImgsAndGt{.345}{JuddsalObjectDB}{00003}{00004}{(c) \textbf{JuddDB}}{.49}\hfill
  \addSampleImgsAndGt{.336}{DutOmron}{img_812503560}{sun_aafvrsnoikyyntky}{(d) \textbf{DUT-OMRON} }{.50}\\ \vspace{.04in}
  \addSampleImgsAndGt{.353}{Thur15k}{Butterfly518}{Giraffe1582}{(e) \textbf{THUR15K}}{.49}\hfill
  \addSampleImgsAndGt{.353}{SED2}{_3076180_cropped2}{yack1}{(f) \textbf{SED2}}{.49}\\ \vspace{-.04in}
  \caption{Images and pixel-level annotations from six salient object datasets.
  }\label{fig:SampleImgs}
\end{figure}

In \figref{fig:dbStats}(c), we demonstrate the average object sizes of these benchmarks,
while the size of each object is normalized by the size of the corresponding image.
We can see that \textbf{MSRA10K} and \textbf{ECCSD} datasets have larger objects while
\textbf{SED2} has smaller ones.
In particular, we can see that some benchmarks contain a limited number of
image regions with large foreground objects.
By jointly considering the center-bias property,
it becomes very easy to achieve a high precision on these images.

\subsection{Evaluation Measures}
There are several ways to measure the agreement between model
predictions and human annotations~\cite{borji2013stands}.
Some metrics evaluate the overlap between a tagged region and and model predictions while others try
to assess the accuracy of drawn shapes with object boundary. In addition, some metrics have tried to consider both boundary and shape~\cite{movahedi2010design}.

Here, we use four universally-agreed, standard, and easy-to-understand measures
for evaluating a salient object detection model.
The first two evaluation metrics are based on the overlapping area between
subjective annotation and saliency prediction,
including the precision-recall (PR) and the receiver operating characteristics (ROC).
From these two metrics, we also report the F-Measure,
which jointly considers recall and precision, and AUC,
which is the area under the ROC curve.
The third measure directly computes the mean absolute
error (MAE) between the estimated saliency map and ground-truth annotation.
For the sake of simplification, we use $S$ to represent the predicted saliency map normalized
to $[0,255]$ and $G$ to represent the ground-truth binary mask of salient objects.
For a binary mask, we use $|\cdot|$ to represent the number of non-zero entries in the mask.
Moreover, we also use the fourth measure proposed by Margolin \etal~\cite{margolinevaluate} which remedies some problems with the classic F-measure for evaluating foreground-background maps obtained using segmentation algorithms.

\myPara{Precision-recall (PR)}
For a saliency map $S$, we can convert it to a binary mask $M$ and compute
$Precision$ and $Recall$ by comparing $M$ with ground-truth $G$:
\begin{align}
    Precision = \frac{|M\cap G|}{|M|}, \ \ Recall = \frac{|M\cap G|}{|G|}
\end{align}\label{eqn:precision_recall}
From this definition, we can see that the binarization of $S$ is the key step in the evaluation.
Usually, there are three popular ways to perform the binarization.
In the first solution, Achanta~\etal~\cite{achanta2009frequency} proposed
the image-dependent adaptive threshold for binarizing $S$,
which is computed as twice as the mean saliency of $S$:
\begin{equation}
    T_{a} = \frac{2}{W \times H} \sum\nolimits_{x=1}^{W}\sum\nolimits_{y=1}^{H} S(x,y)
\end{equation}
where $W$ and $H$ are the width and the height of the saliency map $S$, respectively.

The second way to partition $S$ is to use a fixed threshold
which changes from 0 to 255.
On each threshold, a pair of precision/recall scores are computed,
and are finally combined to form a precision-recall (PR) curve
to describe the model performance at different situations.

The third way of binarization is to use the
SaliencyCut algorithm \cite{ChengPAMI}.
In this solution, a loose threshold,
which typically results in good recall but relatively poor precision,
is used to generate the initial binary mask.
Then the method iteratively uses the GrabCut segmentation method~\cite{RotherKB04Grab}
to gradually refine the binary mask.
The final binary mask is used to re-compute the precision-recall value.

\myPara{F-measure}
Usually, neither $Precision$ nor $Recall$ can comprehensively evaluate
the quality of a saliency map.
To this end, the F-measure is proposed as a weighted harmonic mean of
them with a non-negative weight $\beta$:
\begin{equation}
F_{\beta} = \frac{(1+\beta^{2}) Precision  \times  Recall}{\beta^{2} Precision + Recall}
\label{eq:FMeasure}
\end{equation}
As suggested by many salient object detection works
(\eg, \cite{achanta2009frequency,ChengPAMI, perazzi2012saliency}),
$\beta^{2}$ is set to $0.3$ to increase the importance of the $Precision$ value.
The reason for weighting precision more than recall is that
recall rate is not as important as precision (see also \cite{liu2011learning}).
For instance, $100\%$ recall can be easily achieved by setting the whole region to foreground.

According to the different ways for saliency map binarization,
there exist two ways to compute F-Measure.
When the adaptive threshold or GrabCut algorithm is used for the binarization,
we can generate a single $F_{\beta}$ for each image
and the final F-Measure is computed as the average $F_{\beta}$.
When using fixed thresholding, the resulted PR curve
can be scored by its maximal $F_{\beta}$,
which is a good summary of the detection performance (as suggested in \cite{martin2004learning}).
As defined in \equref{eq:FMeasure},
F-Measure is the weighted harmonic mean of precision and recall,
thus share the same value bounds as precision and recall values, \ie~[0, 1].

\myPara{Receiver operating characteristics (ROC) curve}
In addition to the $Precision$, $Recall$ and $F_{\beta}$,
we can also report the false positive rate ($FPR$) and true positive rate
($TPR$) when binarizing the saliency map with a set of fixed thresholds:
\begin{align}
    TPR = \frac{|M\cap G|}{|G|}, \ \
    FPR = \frac{|M\cap \bar{G}|}{|\bar{G}|}
\end{align}
where $\bar{M}$ and $\bar{G}$ denote the complement of the binary mask
$M$ and ground-truth, respectively.
The ROC curve is the plot of $TPR$ versus $FPR$ by varying the threshold $T_f$.

\myPara{Area under ROC curve (AUC) score}
While ROC is a two-dimensional representation of a model's  performance,
the AUC distills this information into a single scalar.
As the name implies, it is calculated as the area under the ROC curve.
A perfect model will score an AUC of 1,
while random guessing will score an AUC around 0.5.

\begin{figure}[t]
  \centering
  \includegraphics[width=.9\linewidth]{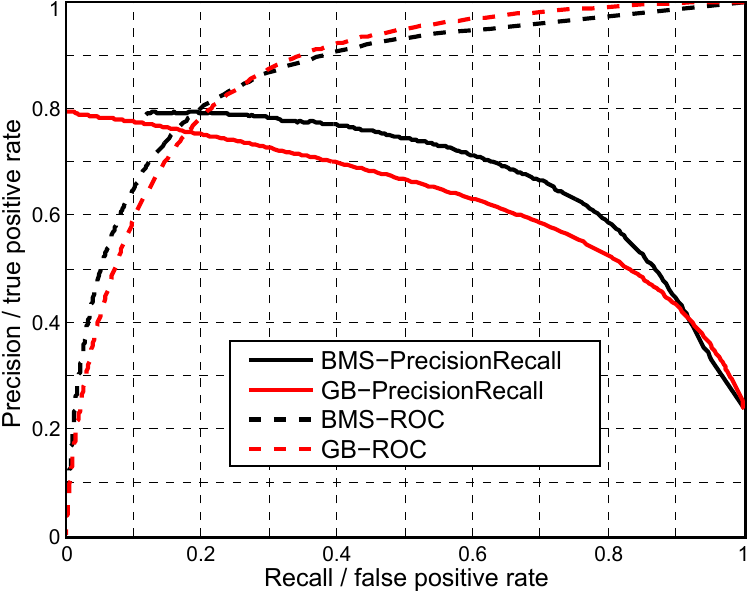}\\
  \caption{PR and ROC curves for BMS~\cite{zhang2013boolean} and
    GB~\cite{harel2007graph} over \textbf{ECSSD}.
  }\label{fig:PrRoc}
\end{figure}

\newcommand{\addFig}[3]{\begin{overpic}[width=.32\linewidth,height=5cm]{ResNew/#2}\put(#3,8){#1}\end{overpic}}
\begin{figure*}[t]
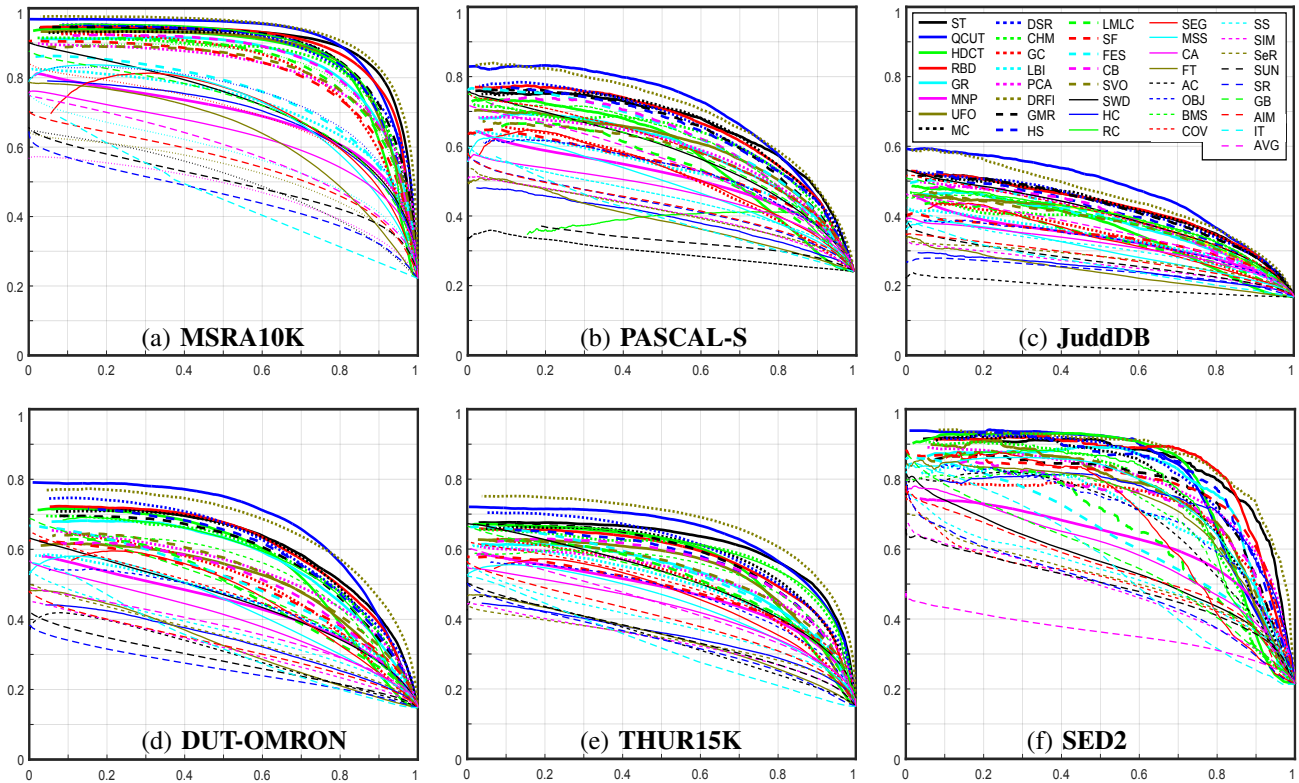

    \centering
    \hspace{-.2in}
    \addFig{(a) \textbf{MSRA10K}}{MSRA10K_Pr}{33} \hspace{.02in}
    \addFig{(b) \textbf{PASCAL-S}}{PASCAL-S_Pr}{33}   \hspace{.02in}
    \addFig{(c) \textbf{JuddDB}}{JuddDB_Pr}{33} \\ \vspace{.15in}
    \hspace{-.2in}
    \addFig{(d) \textbf{DUT-OMRON}}{DUTOMRON_Pr}{33}  \hspace{.02in}
    \addFig{(e) \textbf{THUR15K}}{THUR15K_Pr}{33} \hspace{.02in}
    \addFig{(f) \textbf{SED2}}{SED2_Pr}{35} \\
    \caption{Precision (vertical axis) and recall (horizontal axis) curves of
       saliency methods on 6 popular benchmark datasets.
   }\label{fig:PrCurve}
\end{figure*}

\myPara{Mean absolute error (MAE) score}
The overlap-based evaluation measures introduced above do not
consider the true negative saliency assignments,
\ie, the pixels correctly marked as non-salient.
This favors methods that successfully assign saliency to salient pixels
but fail to detect non-salient regions over methods that successfully
detect non-salient pixels but make mistakes in determining the salient ones
\cite{perazzi2012saliency,ChengWLZVC13Efficient}.
Moreover, in some application scenarios~\cite{avidan2007seam} the quality of the weighted,
continuous saliency maps may be of higher importance than the binary masks.
For a more comprehensive comparison, we therefore also evaluate the mean absolute error
(MAE) between the continuous saliency map $\bar{S}$ and the binary ground truth $\bar{G}$,
both normalized in the range [0, 1]. The MAE score is defined as:

\begin{equation}
MAE = \frac{1}{W \times H} \sum\nolimits_{x=1}^{W}\sum\nolimits_{y=1}^{H} | \bar{S}(x,y) - \bar{G}(x,y)|
\label{MAE}
\end{equation}

\myPara{$\mathbf{F_\beta^w}$-measure} Here, we adopt the technique proposed by Margolin \etal~\cite{margolinevaluate} for quantitative evaluation of models.
As an intuitive generalization of the $F_\beta$-measure,
the new evaluation metric ($F_\beta^w$-measure)  provides reliable evaluation
by i) extending the basic quantities (true-positive, true-negative, false-positive,
and false negative) to non-binary values,
and ii) weighting errors according to their location and their neighborhood.
$F_\beta^w$-measure offers unified solution for evaluation of binary and non-binary maps.

\begin{figure*}[t]
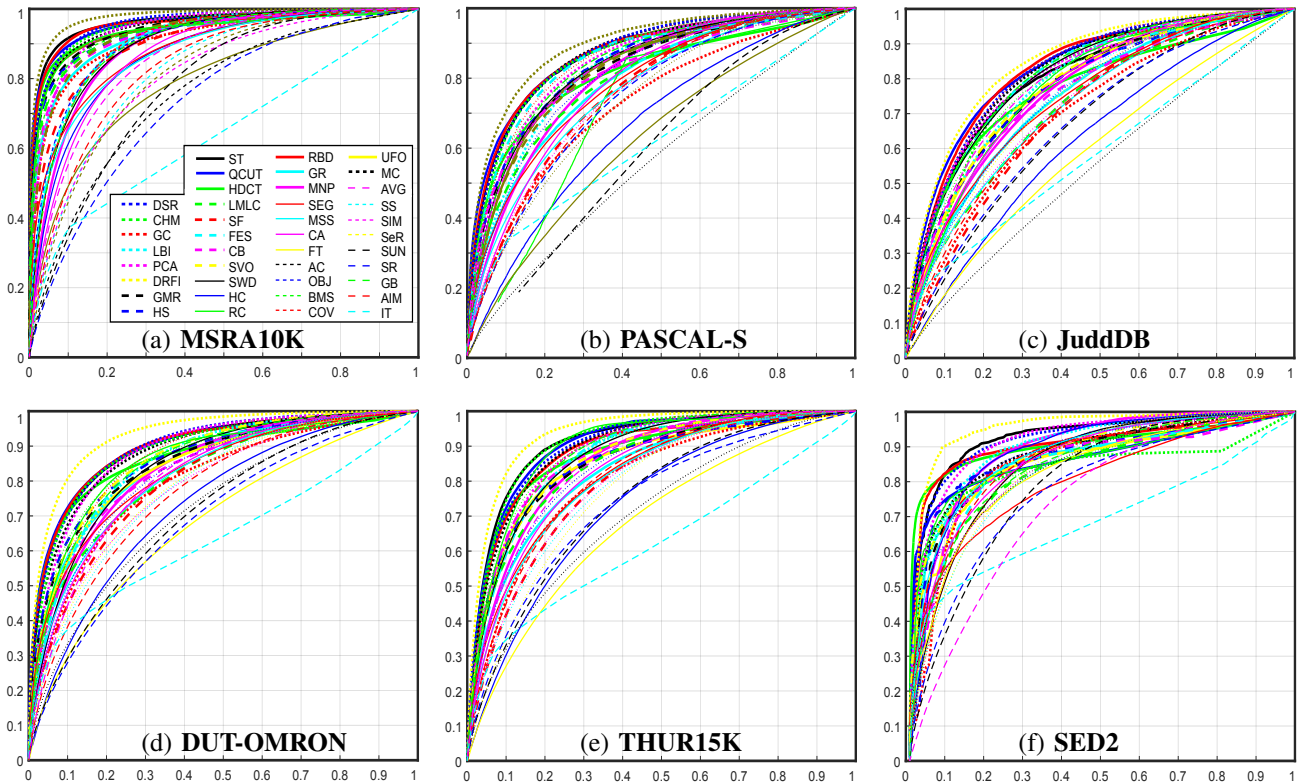

    \centering
    \hspace{-.2in}
    \addFig{(a) \textbf{MSRA10K}}{MSRA10K_Roc}{33} \hspace{.02in}
    \addFig{(b) \textbf{PASCAL-S}}{PASCAL-S_Roc}{33}  \hspace{.02in}
    \addFig{(c) \textbf{JuddDB}}{JuddDB_Roc}{33}  \\  \vspace{.15in}
    \hspace{-.2in}
    \addFig{(d) \textbf{DUT-OMRON}}{DUTOMRON_Roc}{33} \hspace{.02in}
    \addFig{(e) \textbf{THUR15K}}{THUR15K_Roc}{33} \hspace{.02in}
    \addFig{(f) \textbf{SED2}}{SED2_Roc}{33} \\
    \caption{ROC curves of models on 6 benchmarks.
        False and true positive rates are shown in $x$ and $y$ axes, respectively.
   }\label{fig:RocCurve}
\end{figure*}

\begin{figure*}[t]
\renewcommand{\tabTitle}{ \textbf{Model}&\scriptsize {PASCAL}&\scriptsize {THUR}&\scriptsize {JuddDB}&\scriptsize {DUT}&\scriptsize {MSRA}&\scriptsize {ECSSD}&\scriptsize {SED2}}
\begin{minipage}[t]{0.48\linewidth}
    \centering \small
    \renewcommand{\arraystretch}{1.1}
    \renewcommand{\tabcolsep}{1.25mm}
    \begin{tabular}{|l||c|c|c|c|c|c|c|} \hline
	\tabTitle \\	\hline \hline
	\textbf{ST}   & .868 & \second{.911} & .806 & .895 & \second{.961} & \third{.914} & \second{.922} \\
	\textbf{QCUT} & .870 & .907 & \second{.831} & \third{.897} & .956 & .909 & .860 \\
	\textbf{HDCT} & .815 & .878 & .771 & .869 & .941 & .866 & .898 \\
	\textbf{RBD}  & .867 & .887 & \third{.826} & .894 & .955 & .894 & .899 \\
	\textbf{GR}   & .794 & .829 & .747 & .846 & .925 & .831 & .854 \\
	\textbf{MNP}  & .807 & .854 & .768 & .835 & .895 & .820 & .888 \\
	\textbf{UFO}  & .825 & .853 & .775 & .839 & .938 & .875 & .845 \\
	\textbf{MC}   & \second{.873} & .895 & .823 & .887 & .951 & .910 & .877 \\
	\textbf{DSR}  & \third{.873} & .902 & .826 & \second{.899} & \third{.959} & \second{.914} & \third{.915} \\
	\textbf{CHM}  & .864 & \third{.910} & .797 & .890 & .952 & .903 & .831 \\
	\textbf{GC}   & .728 & .803 & .702 & .796 & .912 & .805 & .846 \\
	\textbf{LBI}  & .828 & .876 & .792 & .854 & .910 & .842 & .896 \\
	\textbf{PCA}  & .848 & .885 & .804 & .887 & .941 & .876 & .911 \\
	\textbf{DRFI} & \first{.897} & \first{.938} & \first{.851} & \first{.933} & \first{.978} & \first{.944} & \first{.944} \\
	\textbf{GMR}  & .829 & .856 & .781 & .853 & .944 & .889 & .862 \\
	\textbf{HS}   & .840 & .853 & .775 & .860 & .933 & .883 & .858 \\
	\textbf{LMLC} & .800 & .853 & .724 & .817 & .936 & .849 & .826 \\
	\textbf{SF}   & .762 & .799 & .711 & .803 & .905 & .817 & .871 \\
	\textbf{FES}  & .854 & .867 & .805 & .848 & .898 & .860 & .838 \\
	\textbf{CB}   & .818 & .870 & .760 & .831 & .927 & .875 & .839 \\
	\textbf{SVO}  & .826 & .865 & .784 & .866 & .930 & .857 & .875 \\
	\textbf{SWD}  & .835 & .873 & .812 & .843 & .901 & .857 & .845 \\
	\textbf{HC}   & .669 & .735 & .626 & .733 & .867 & .704 & .880 \\
	\textbf{RC}   & .713 & .896 & .775 & .859 & .936 & .892 & .852 \\
	\textbf{SEG}  & .787 & .818 & .747 & .825 & .882 & .808 & .796 \\
	\textbf{MSS}  & .766 & .813 & .726 & .817 & .875 & .779 & .871 \\
	\textbf{CA}   & .782 & .830 & .774 & .815 & .872 & .784 & .853 \\
	\textbf{FT}   & .629 & .684 & .593 & .682 & .790 & .661 & .820 \\
	\textbf{AC}   & .565 & .707 & .548 & .721 & .756 & .668 & .831 \\
	\hline \hline
	\textbf{OBJ}  & .811 & .839 & .750 & .822 & .907 & .818 & .870 \\
	\hline \hline
	\textbf{BMS}  & .834 & .879 & .788 & .856 & .929 & .865 & .852 \\
	\textbf{COV}  & .856 & .883 & .826 & .864 & .904 & .879 & .833 \\
	\textbf{SS}   & .752 & .792 & .754 & .784 & .823 & .725 & .826 \\
	\textbf{SIM}  & .756 & .797 & .727 & .783 & .808 & .734 & .833 \\
	\textbf{SeR}  & .736 & .778 & .746 & .786 & .813 & .695 & .835 \\
	\textbf{SUN}  & .588 & .746 & .674 & .708 & .778 & .623 & .789 \\
	\textbf{SR}   & .745 & .741 & .676 & .688 & .736 & .633 & .769 \\
	\textbf{GB}   & .843 & .882 & .815 & .857 & .902 & .865 & .839 \\
	\textbf{AIM}  & .753 & .814 & .719 & .768 & .833 & .730 & .846 \\
	\textbf{IT}   & .621 & .623 & .586 & .636 & .640 & .577 & .682 \\
	\hline \hline
	\textbf{AAM}  & .835 & .849 & .797 & .814 & .857 & .863 & .736 \\
\hline
\end{tabular}

    \caption{AUC: area under ROC curve (Higher is better. The top three models are highlighted in red, green and blue). }\label{fig:AUC}
\end{minipage}
\hspace{0.15cm}
\begin{minipage}[t]{0.48\linewidth}
    \centering \small
    \renewcommand{\arraystretch}{1.1}
    \renewcommand{\tabcolsep}{1.25mm}
    \begin{tabular}{|l||c|c|c|c|c|c|c|} \hline
	\tabTitle \\	\hline \hline
	\textbf{ST}   & .224 & .179 & .240 & .182 & .122 & .193 & .145 \\
	\textbf{QCUT} & \first{.195} & \first{.128} & \first{.178} & \first{.119} & \second{.118} & \second{.171} & .148 \\
	\textbf{HDCT} & .229 & .177 & .209 & .164 & .143 & .199 & .162 \\
	\textbf{RBD}  & \second{.199} & .150 & .212 & \third{.144} & \first{.108} & .173 & \second{.130} \\
	\textbf{GR}   & .299 & .256 & .311 & .259 & .198 & .285 & .189 \\
	\textbf{MNP}  & .298 & .255 & .286 & .272 & .229 & .307 & .215 \\
	\textbf{UFO}  & .245 & .165 & .216 & .173 & .150 & .207 & .180 \\
	\textbf{MC}   & .230 & .184 & .231 & .186 & .145 & .204 & .182 \\
	\textbf{DSR}  & \third{.204} & \second{.142} & .196 & \second{.139} & .121 & \third{.173} & \third{.140} \\
	\textbf{CHM}  & .222 & .153 & .226 & .152 & .142 & .195 & .168 \\
	\textbf{GC}   & .265 & .192 & .258 & .197 & .139 & .214 & .185 \\
	\textbf{LBI}  & .281 & .239 & .273 & .249 & .224 & .280 & .207 \\
	\textbf{PCA}  & .245 & .198 & \second{.181} & .206 & .185 & .248 & .200 \\
	\textbf{DRFI} & .221 & \third{.150} & .213 & .155 & \third{.118} & \first{.166} & \first{.130} \\
	\textbf{GMR}  & .233 & .181 & .243 & .189 & .126 & .189 & .163 \\
	\textbf{HS}   & .262 & .218 & .282 & .227 & .149 & .228 & .157 \\
	\textbf{LMLC} & .284 & .246 & .303 & .277 & .163 & .260 & .269 \\
	\textbf{SF}   & .253 & .184 & .218 & .183 & .175 & .230 & .180 \\
	\textbf{FES}  & .218 & .155 & .184 & .156 & .185 & .215 & .196 \\
	\textbf{CB}   & .272 & .227 & .287 & .257 & .178 & .241 & .195 \\
	\textbf{SVO}  & .392 & .382 & .422 & .409 & .331 & .404 & .348 \\
	\textbf{SWD}  & .315 & .288 & .292 & .310 & .267 & .318 & .296 \\
	\textbf{HC}   & .354 & .291 & .348 & .310 & .215 & .331 & .193 \\
	\textbf{RC}   & .304 & .168 & .270 & .189 & .137 & .187 & .148 \\
	\textbf{SEG}  & .350 & .336 & .354 & .337 & .298 & .342 & .312 \\
	\textbf{MSS}  & .251 & .178 & .204 & .177 & .203 & .245 & .192 \\
	\textbf{CA}   & .299 & .248 & .282 & .254 & .237 & .310 & .229 \\
	\textbf{FT}   & .307 & .241 & .267 & .250 & .235 & .291 & .206 \\
	\textbf{AC}   & .286 & .195 & .239 & .190 & .227 & .265 & .206 \\
	\hline \hline
	\textbf{OBJ}  & .334 & .306 & .359 & .323 & .262 & .337 & .269 \\
	\hline \hline
	\textbf{BMS}  & .225 & .181 & .233 & .175 & .151 & .216 & .184 \\
	\textbf{COV}  & .226 & .155 & \third{.182} & .156 & .197 & .217 & .210 \\
	\textbf{SS}   & .315 & .267 & .301 & .277 & .266 & .344 & .266 \\
	\textbf{SIM}  & .414 & .414 & .412 & .429 & .388 & .433 & .384 \\
	\textbf{SeR}  & .371 & .345 & .379 & .352 & .310 & .404 & .290 \\
	\textbf{SUN}  & .467 & .310 & .319 & .349 & .306 & .396 & .307 \\
	\textbf{SR}   & .291 & .175 & .200 & .181 & .232 & .266 & .220 \\
	\textbf{GB}   & .270 & .229 & .261 & .240 & .222 & .263 & .242 \\
	\textbf{AIM}  & .337 & .298 & .331 & .322 & .286 & .339 & .262 \\
	\textbf{IT}   & .240 & .199 & .200 & .198 & .213 & .273 & .245 \\
	\hline \hline
	\textbf{AAM}  & .316 & .248 & .343 & .288 & .260 & .276 & .405 \\
\hline
\end{tabular}

    \caption{MAE: Mean Absolute Error (Smaller is better.  The top three models are highlighted in red, green and blue).}\label{fig:MAE}
\end{minipage}
\end{figure*}

\renewcommand{\tabTitle}{ \textbf{Model}&  {PASCAL}&  {THUR}&  {JuddDB}&  {DUT}&  {MSRA}&  {ECSSD}&  {SED2}}
\begin{figure}[t]
    \fontsize{8.8}{1em}\selectfont
    \begin{tabular}{|l||c|c|c|c|c|c|c|} \hline
	\tabTitle \\	\hline \hline
	\textbf{ST}   & \first{.476} & .415 & \third{.274} & .393 & \second{.660} & \third{.491} & \third{.585} \\
	\textbf{QCUT} & .383 & .421 & \second{.293} & \first{.448} & .627 & .467 & .533 \\
	\textbf{HDCT} & .384 & .360 & .233 & .365 & .582 & .430 & .520 \\
	\textbf{RBD}  & \second{.472} & \third{.421} & \first{.308} & \second{.429} & \first{.685} & .490 & \first{.642} \\
	\textbf{GR}   & .371 & .276 & .211 & .286 & .485 & .355 & .487 \\
	\textbf{MNP}  & .370 & .274 & .199 & .270 & .416 & .335 & .417 \\
	\textbf{UFO}  & .381 & .353 & .231 & .334 & .556 & .405 & .477 \\
	\textbf{MC}   & .422 & .349 & .256 & .355 & .576 & .441 & .465 \\
	\textbf{DSR}  & .439 & \second{.422} & .273 & \third{.420} & \third{.656} & .490 & .583 \\
	\textbf{CHM}  & .391 & .378 & .235 & .366 & .573 & .424 & .463 \\
	\textbf{GC}   & .396 & .367 & .244 & .358 & .612 & .437 & .555 \\
	\textbf{LBI}  & .373 & .286 & .210 & .276 & .443 & .346 & .435 \\
	\textbf{PCA}  & .353 & .298 & .205 & .303 & .473 & .358 & .437 \\
	\textbf{DRFI} & .421 & \first{.444} & .264 & .417 & .654 & \first{.516} & \second{.638} \\
	\textbf{GMR}  & .438 & .380 & .274 & .384 & .643 & .476 & .584 \\
	\textbf{HS}   & \third{.451} & .365 & .251 & .356 & .604 & .449 & .572 \\
	\textbf{LMLC} & .439 & .343 & .236 & .313 & .594 & .428 & .399 \\
	\textbf{SF}   & .280 & .259 & .168 & .279 & .440 & .309 & .448 \\
	\textbf{FES}  & .306 & .286 & .202 & .276 & .388 & .312 & .267 \\
	\textbf{CB}   & .391 & .318 & .232 & .284 & .528 & .403 & .463 \\
	\textbf{SVO}  & .296 & .229 & .206 & .228 & .363 & .329 & .323 \\
	\textbf{SWD}  & .355 & .244 & .198 & .239 & .365 & .332 & .317 \\
	\textbf{HC}   & .343 & .255 & .186 & .243 & .481 & .323 & .507 \\
	\textbf{RC}   & .326 & .414 & .254 & .377 & .608 & \second{.493} & .567 \\
	\textbf{SEG}  & .349 & .223 & .194 & .234 & .332 & .323 & .294 \\
	\textbf{MSS}  & .241 & .229 & .144 & .225 & .345 & .251 & .382 \\
	\textbf{CA}   & .333 & .244 & .192 & .243 & .379 & .304 & .348 \\
	\textbf{FT}   & .241 & .195 & .135 & .191 & .334 & .239 & .393 \\
	\textbf{AC}   & .167 & .178 & .106 & .166 & .172 & .191 & .323 \\
	\hline \hline
	\textbf{OBJ}  & .373 & .252 & .207 & .250 & .403 & .339 & .383 \\
	\hline \hline
	\textbf{BMS}  & .404 & .348 & .233 & .358 & .566 & .420 & .447 \\
	\textbf{COV}  & .250 & .252 & .186 & .254 & .303 & .277 & .256 \\
	\textbf{SS}   & .294 & .204 & .177 & .210 & .312 & .268 & .309 \\
	\textbf{SIM}  & .338 & .195 & .176 & .201 & .293 & .291 & .294 \\
	\textbf{SeR}  & .354 & .225 & .198 & .229 & .352 & .295 & .359 \\
	\textbf{SUN}  & .310 & .194 & .156 & .188 & .301 & .254 & .308 \\
	\textbf{SR}   & .277 & .134 & .091 & .118 & .155 & .138 & .184 \\
	\textbf{GB}   & .354 & .269 & .210 & .266 & .392 & .344 & .335 \\
	\textbf{AIM}  & .307 & .229 & .178 & .216 & .341 & .285 & .355 \\
	\textbf{IT}   & .181 & .119 & .078 & .122 & .228 & .141 & .165 \\
	\hline \hline
	\textbf{AAM}  & .374 & .258 & .219 & .255 & .381 & .365 & .267 \\
\hline
\end{tabular}

    \caption{Evaluation results using $F_\beta^w$-measure \cite{margolinevaluate}.
    }\label{fig:FE}
\end{figure}

\newcommand{\fMeasureTypes}{Fixed & AdpT & SCut}
\renewcommand{\tabTitle}{{\textbf{Model}}&\Cols{\textbf{PASCAL-S}}&\Cols{\textbf{THUR15K}}&\Cols{\textbf{JuddDB}}&\Cols{\textbf{DUT-OMRON}}
   &\Cols{\textbf{MSRA10K}}&\Cols{\textbf{ECSSD}}&\Cols{\textbf{SED2}}\\ \cline{2-22}
   &\fMeasureTypes & \fMeasureTypes & \fMeasureTypes & \fMeasureTypes & \fMeasureTypes & \fMeasureTypes & \fMeasureTypes}

\begin{figure*}[t]
\begin{minipage}[t]{\linewidth}
    \centering
    \small
    \renewcommand{\arraystretch}{1.1}
    \renewcommand{\tabcolsep}{0.32mm}
    \begin{tabular}{|l||ccc|ccc|ccc|ccc|ccc|ccc|ccc|} \hline
	\tabTitle \\	\hline \hline
	\textbf{ST}   & .660 & .601 & \second{.671} & \third{.631} & .580 & .648 & .455 & .394 & \third{.459} & \third{.631} & .577 & .635 & \third{.868} & .825 & \second{.896} & \third{.752} & .690 & \third{.777} & .818 & .805 & \first{.768} \\
	\textbf{QCUT} & \first{.695} & \first{.654} & .613 & \second{.651} & \first{.625} & .620 & \first{.509} & \first{.454} & \first{.480} & \first{.683} & \first{.647} & \third{.647} & \second{.874} & \first{.843} & .843 & \second{.779} & \first{.738} & .747 & .810 & .801 & .672 \\
	\textbf{HDCT} & .604 & .572 & .611 & .602 & .571 & .636 & .412 & .378 & .422 & .609 & .572 & .643 & .837 & .807 & .877 & .705 & .669 & .740 & \third{.822} & .802 & .758 \\
	\textbf{RBD}  & .652 & .607 & .667 & .596 & .566 & .618 & .457 & .403 & \second{.461} & .630 & .580 & \second{.647} & .856 & .821 & \third{.884} & .718 & .680 & .757 & \first{.837} & \second{.825} & .750 \\
	\textbf{GR}   & .596 & .508 & .604 & .551 & .509 & .546 & .418 & .338 & .378 & .599 & .540 & .580 & .816 & .770 & .830 & .664 & .583 & .677 & .798 & .753 & .639 \\
	\textbf{MNP}  & .522 & .510 & .630 & .495 & .523 & .603 & .367 & .337 & .405 & .467 & .486 & .576 & .668 & .724 & .822 & .568 & .555 & .709 & .621 & .778 & \second{.765} \\
	\textbf{UFO}  & .606 & .554 & .622 & .579 & .557 & .610 & .432 & .385 & .433 & .545 & .541 & .593 & .842 & .806 & .862 & .701 & .654 & .739 & .742 & .781 & .729 \\
	\textbf{MC}   & \third{.661} & \second{.622} & \third{.670} & .610 & .603 & .600 & \third{.460} & .420 & .434 & .627 & .603 & .615 & .847 & .824 & .855 & .742 & .704 & .745 & .779 & .803 & .630 \\
	\textbf{DSR}  & .646 & \third{.619} & .650 & .611 & \third{.604} & .597 & .454 & \third{.421} & .410 & .626 & \second{.614} & .593 & .835 & .824 & .833 & .737 & \third{.717} & .703 & .794 & \third{.821} & .632 \\
	\textbf{CHM}  & .631 & .586 & .634 & .612 & .591 & .643 & .417 & .368 & .424 & .604 & .586 & .637 & .825 & .804 & .857 & .722 & .684 & .735 & .750 & .750 & .658 \\
	\textbf{GC}   & .535 & .472 & .553 & .533 & .517 & .497 & .384 & .321 & .342 & .535 & .528 & .506 & .794 & .777 & .780 & .641 & .612 & .593 & .729 & .730 & .616 \\
	\textbf{LBI}  & .538 & .525 & .629 & .519 & .534 & .618 & .371 & .353 & .416 & .482 & .504 & .609 & .696 & .714 & .857 & .586 & .563 & .738 & .692 & .776 & \third{.764} \\
	\textbf{PCA}  & .593 & .567 & .634 & .544 & .558 & .601 & .432 & .404 & .368 & .554 & .554 & .624 & .782 & .782 & .845 & .646 & .627 & .720 & .754 & .796 & .701 \\
	\textbf{DRFI} & \second{.679} & .615 & \first{.690} & \first{.670} & \second{.607} & \first{.674} & \second{.475} & .419 & .447 & \second{.665} & \third{.605} & \first{.669} & \first{.881} & \second{.838} & \first{.905} & \first{.787} & \second{.733} & \first{.801} & \second{.831} & \first{.839} & .702 \\
	\textbf{GMR}  & .643 & .607 & .654 & .597 & .594 & .579 & .454 & .409 & .432 & .610 & .591 & .591 & .847 & \third{.825} & .839 & .740 & .712 & .736 & .773 & .789 & .643 \\
	\textbf{HS}   & .637 & .559 & .647 & .585 & .549 & .602 & .442 & .358 & .428 & .616 & .565 & .616 & .845 & .800 & .870 & .731 & .659 & .769 & .811 & .776 & .713 \\
	\textbf{LMLC} & .555 & .505 & .614 & .540 & .519 & .588 & .375 & .302 & .397 & .521 & .493 & .551 & .801 & .772 & .860 & .659 & .600 & .735 & .653 & .712 & .674 \\
	\textbf{SF}   & .544 & .488 & .461 & .500 & .495 & .342 & .373 & .319 & .219 & .519 & .512 & .377 & .779 & .759 & .573 & .619 & .576 & .378 & .764 & .794 & .509 \\
	\textbf{FES}  & .619 & .605 & .534 & .547 & .575 & .426 & .424 & .411 & .333 & .520 & .555 & .380 & .717 & .753 & .534 & .645 & .655 & .467 & .617 & .785 & .174 \\
	\textbf{CB}   & .623 & .561 & .636 & .581 & .556 & .615 & .444 & .375 & .435 & .542 & .534 & .593 & .815 & .775 & .857 & .717 & .656 & .761 & .730 & .704 & .657 \\
	\textbf{SVO}  & .586 & .361 & .621 & .554 & .441 & .609 & .414 & .279 & .419 & .557 & .407 & .609 & .789 & .585 & .863 & .639 & .357 & .737 & .744 & .667 & .746 \\
	\textbf{SWD}  & .577 & .523 & .642 & .528 & .560 & \third{.649} & .434 & .386 & .454 & .478 & .506 & .613 & .689 & .705 & .871 & .624 & .549 & \second{.781} & .548 & .714 & .737 \\
	\textbf{HC}   & .423 & .383 & .464 & .386 & .401 & .436 & .286 & .257 & .280 & .382 & .380 & .435 & .677 & .663 & .740 & .460 & .441 & .499 & .736 & .759 & .646 \\
	\textbf{RC}   & .466 & .351 & .470 & .610 & .586 & .639 & .431 & .370 & .425 & .599 & .578 & .621 & .844 & .820 & .875 & .741 & .701 & .776 & .774 & .807 & .649 \\
	\textbf{SEG}  & .534 & .344 & .627 & .500 & .425 & .580 & .376 & .268 & .393 & .516 & .450 & .562 & .697 & .585 & .812 & .568 & .408 & .715 & .704 & .640 & .669 \\
	\textbf{MSS}  & .503 & .485 & .399 & .478 & .490 & .200 & .341 & .324 & .089 & .476 & .490 & .193 & .696 & .711 & .362 & .530 & .536 & .203 & .743 & .783 & .298 \\
	\textbf{CA}   & .489 & .472 & .586 & .458 & .494 & .557 & .353 & .330 & .394 & .435 & .458 & .532 & .621 & .679 & .748 & .515 & .494 & .625 & .591 & .737 & .565 \\
	\textbf{FT}   & .408 & .367 & .357 & .386 & .400 & .238 & .278 & .250 & .132 & .381 & .388 & .259 & .635 & .628 & .472 & .434 & .431 & .257 & .715 & .734 & .436 \\
	\textbf{AC}   & .326 & .279 & .265 & .382 & .431 & .068 & .227 & .199 & .049 & .354 & .383 & .040 & .520 & .566 & .014 & .411 & .410 & .038 & .684 & .729 & .140 \\
	\hline \hline
	\textbf{OBJ}  & .544 & .444 & .596 & .498 & .482 & .593 & .368 & .282 & .413 & .481 & .445 & .578 & .718 & .681 & .840 & .574 & .456 & .698 & .685 & .723 & .731 \\
	\hline \hline
	\textbf{BMS}  & .617 & .596 & .624 & .568 & .578 & .594 & .434 & .404 & .416 & .573 & .576 & .580 & .805 & .798 & .822 & .683 & .659 & .690 & .713 & .760 & .627 \\
	\textbf{COV}  & .589 & .604 & .535 & .510 & .587 & .398 & .429 & \second{.427} & .315 & .486 & .579 & .373 & .667 & .755 & .394 & .641 & .677 & .413 & .518 & .724 & .212 \\
	\textbf{SS}   & .469 & .451 & .552 & .415 & .482 & .523 & .344 & .321 & .397 & .396 & .443 & .502 & .572 & .642 & .675 & .467 & .441 & .574 & .533 & .696 & .641 \\
	\textbf{SIM}  & .434 & .407 & .599 & .372 & .429 & .568 & .295 & .292 & .384 & .358 & .402 & .539 & .498 & .585 & .794 & .433 & .391 & .672 & .498 & .685 & .725 \\
	\textbf{SeR}  & .433 & .406 & .566 & .374 & .419 & .536 & .316 & .285 & .388 & .385 & .411 & .532 & .542 & .607 & .755 & .419 & .391 & .596 & .521 & .714 & .702 \\
	\textbf{SUN}  & .359 & .294 & .467 & .387 & .432 & .486 & .303 & .291 & .285 & .321 & .360 & .445 & .505 & .596 & .670 & .388 & .376 & .478 & .504 & .661 & .613 \\
	\textbf{SR}   & .447 & .442 & .497 & .374 & .457 & .002 & .279 & .270 & .001 & .298 & .363 & .000 & .473 & .569 & .001 & .381 & .385 & .001 & .504 & .700 & .002 \\
	\textbf{GB}   & .581 & .567 & .651 & .526 & .571 & \second{.650} & .419 & .396 & .455 & .507 & .548 & .638 & .688 & .737 & .837 & .624 & .613 & .765 & .571 & .746 & .695 \\
	\textbf{AIM}  & .450 & .375 & .593 & .427 & .461 & .559 & .317 & .260 & .360 & .361 & .377 & .495 & .555 & .575 & .750 & .449 & .357 & .571 & .541 & .718 & .693 \\
	\textbf{IT}   & .414 & .453 & .255 & .373 & .437 & .005 & .297 & .283 & .000 & .378 & .449 & .005 & .471 & .586 & .158 & .407 & .414 & .003 & .579 & .697 & .008 \\
	\hline \hline
	\textbf{AAM}  & .549 & .536 & .578 & .458 & .569 & .620 & .392 & .367 & .411 & .406 & .514 & .534 & .580 & .692 & .779 & .597 & .627 & .756 & .388 & .524 & .640 \\
\hline
\end{tabular}

    \caption{$F_\beta$ statistics on each dataset,
        using varying fixed thresholds,
        adaptive threshold, and SaliencyCut (Higher is better. The top three models are highlighted in red, green and blue).
    }\label{fig:FMeasure}
\end{minipage}
\end{figure*}

Note that these scores sometimes do not agree with each other.
For example, \figref{fig:PrRoc} shows a comparison of two models over the
\textbf{ECSSD} dataset using PR and ROC metrics.
While there is not a big difference in ROC curves (thus about the same AUC),
one model clearly scores better using the PR curve (thus having higher $F_\beta$).
Such disparity between the ROC and PR measures has been extensively
studied in \cite{davis2006relationship}.
Note that the number of negative examples (non-salient pixels) is typically much bigger
than the number of positive examples (salient object pixels) in evaluating salient object detection models. Therefore,
PR curves are more informative than ROC curves and can present an over optimistic view of an algorithm's performance
\cite{davis2006relationship}.
Thus we mainly base our conclusions on the PR curves scores (\ie, F-Measure scores),
and also report other scores for comprehensive comparisons
and for facilitating specific application requirements.
It is worth mentioning that active research is ongoing to figure out the better ways
of evaluating salient object detection and segmentation models
(\eg~\cite{margolinevaluate}).

\subsection{Quantitative Comparison of Models}
\label{sec:QuantitativeComparison}

We evaluate saliency maps produced by different models on seven datasets by using all evaluation metrics:
\begin{enumerate}
\item \figref{fig:PrCurve} and~\figref{fig:RocCurve} show PR and ROC curves;

\item \figref{fig:AUC} and~\figref{fig:MAE} demonstrate AUC and MAE scores;

\item \tabref{fig:FE} and \tabref{fig:FMeasure} show $F_\beta^w$ and $F_\beta$ scores of all models, respectively\footnote{Three segmentation methods are used, including adaptive threshold, fixed threshold, and SaliencyCut algorithm. The influence of segmentation methods will be discussed in Sect.~\ref{sec:SalCut}}.

\end{enumerate}

In terms of both \textbf{PR} and \textbf{ROC} curves,
DRFI model surprisingly outperforms all other models on seven benchmark
datasets with large margins.
Besides, RBD, DSR and MC (solid lines with blue, yellow, and magenta colors, respectively)
achieve close performance and perform slightly better than other models.

Using the \textbf{F-measure} (\ie, $F_\beta$), the five best models are: DRFI, MC, RBD, DSR, and GMR, where DRFI model consistently wins over all the 5 datasets.
MC ranks the second best over 2 datasets and the third best over 2 datasets. SR and SIM models perform the worst.

With respect to the \textbf{AUC} score, DRFI again ranks the best over all seven datasets.
Following DRFI, DSR model ranks the second over 4 datasets.
RBD ranks the second on 1 dataset and the third on 2 datasets.
While PCA ranks the third on 1 dataset in terms of AUC score, it is not on the list of top three contenders using $F_\beta$ measure.
IT, SR, and SUN achieve the worst performance.
It is worth being mentioned that all the models perform well above chance level (AUC = 0.5) on seven benchmark datasets.

Rankings of models using \textbf{MAE} are more diverse than either $F_\beta$ or AUC scores.
DSR, RBD and DRFI rank on the top, but none of them are among top three models over \textbf{JuddDB}.
MC, which performs well in terms of $F_\beta$ and AUC, is not included in the top three models on any dataset.
PCA performs the best on \textbf{JuddDB} but worse on others.
SIM and SVO models perform the worst.

Using the \myPara{$\mathbf{F_\beta^w}$-measure}, RBD, DRFI, and ST rank at the top. Other top contenders here are: DSR, QCUT, RC and HS. RBD model ranks better using this score than the other ones.

\begin{figure*}[htbp!]
    \centering	
    \includegraphics[width=.9\linewidth]{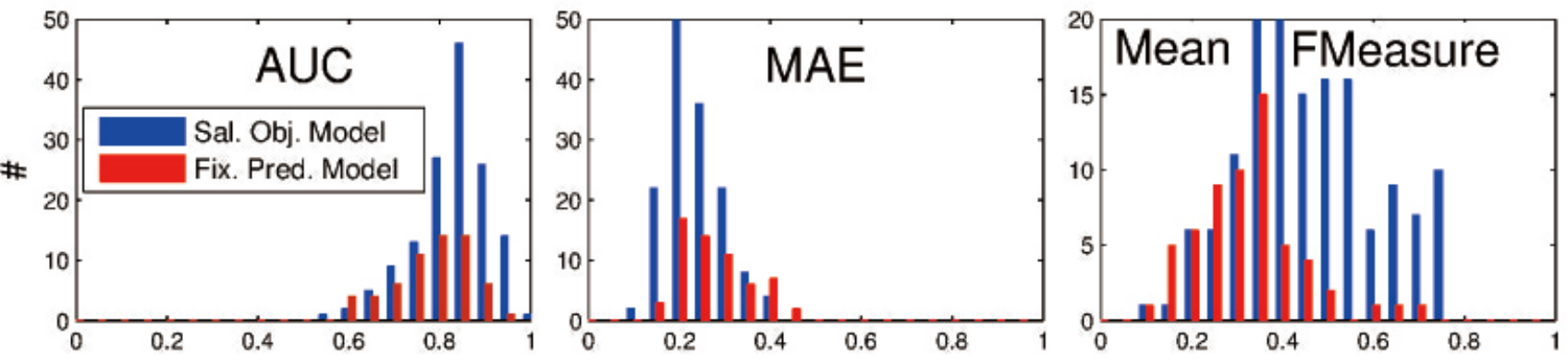} \\
    \caption{Histogram of AUC, MAE, and Mean $F_\beta$ scores for
        salient object detection models (blue) versus fixation prediction
        models (red) collapsed over all datasets.
    }\label{fig:histScores}
\end{figure*}

On average, the compared fixation prediction and object proposal generation models perform worse than salient object detection models.
As two outliers, COV and BMS outperform several salient object detection models in terms of all evaluation metrics, implying that they are suitable for detecting salient proto objects.
Additionally, \figref{fig:histScores} shows the distribution of $F_\beta$,
ROC and MAE scores of all salient object detection models versus
all fixation prediction models over all benchmark datasets.
We can see a sharp separation of models especially for the $F_\beta$ score,
where most of the top models are salient object detection models.
This result is consistent with the conclusion in~\cite{borji2012salient} that fixation prediction models perform lower than salient object
detection models.
Though stemming from fixation prediction,
research in salient object detection shares its unique properties and has truly
added to what traditional saliency models focusing on fixation prediction already offer.

In particular, most of the salient object detection models outperform the baseline AAM model. Among these 29 models,
AAM only outperforms 1 model over \textbf{MSRA10K}, 8 models over \textbf{ECSSD}, 3 on \textbf{THUR15K}, 11 on \textbf{JuddDB}, 9 on \textbf{PASCAL-S} and 3 on \textbf{DUT-OMRON} in terms of $F_\beta$ (Fixed).
Interestingly, AAM model does not outperform any model over \textbf{SED2}, which means that indeed there is less center bias in this dataset
and salient object detection models can detect off-center objects.
Notice that AAM ranks lowest on \textbf{SED2} compared to other datasets.
Please notice that it does not necessarily mean that models below AAM are not good, as taking advantage of the location prior may further
enhance their performance (\eg, AC and FT).

On average, over all models and scores, the performances were lower on \textbf{JuddDB},
\textbf{PASCAL-S} and \textbf{THUR15K}, implying that these datasets were more challenging.
The low model performance of \textbf{JuddDB} can be caused by both less
center bias and small objects in images.
By investigating some images of these datasets for which models performed low,
we found that there are several objects that can be potentially
the most salient one.
This makes the generation of ground-truth quite subjective and challenging,
although the most salient object in \textbf{JuddDB} and \textbf{PASCAL-S} datasets has objectively been defined to
be the most looked-at object measured from eye movement data.



\subsection{Qualitative Comparison of Models}

\begin{figure}[t]
    \centering	
    \includegraphics[width=0.9\linewidth]{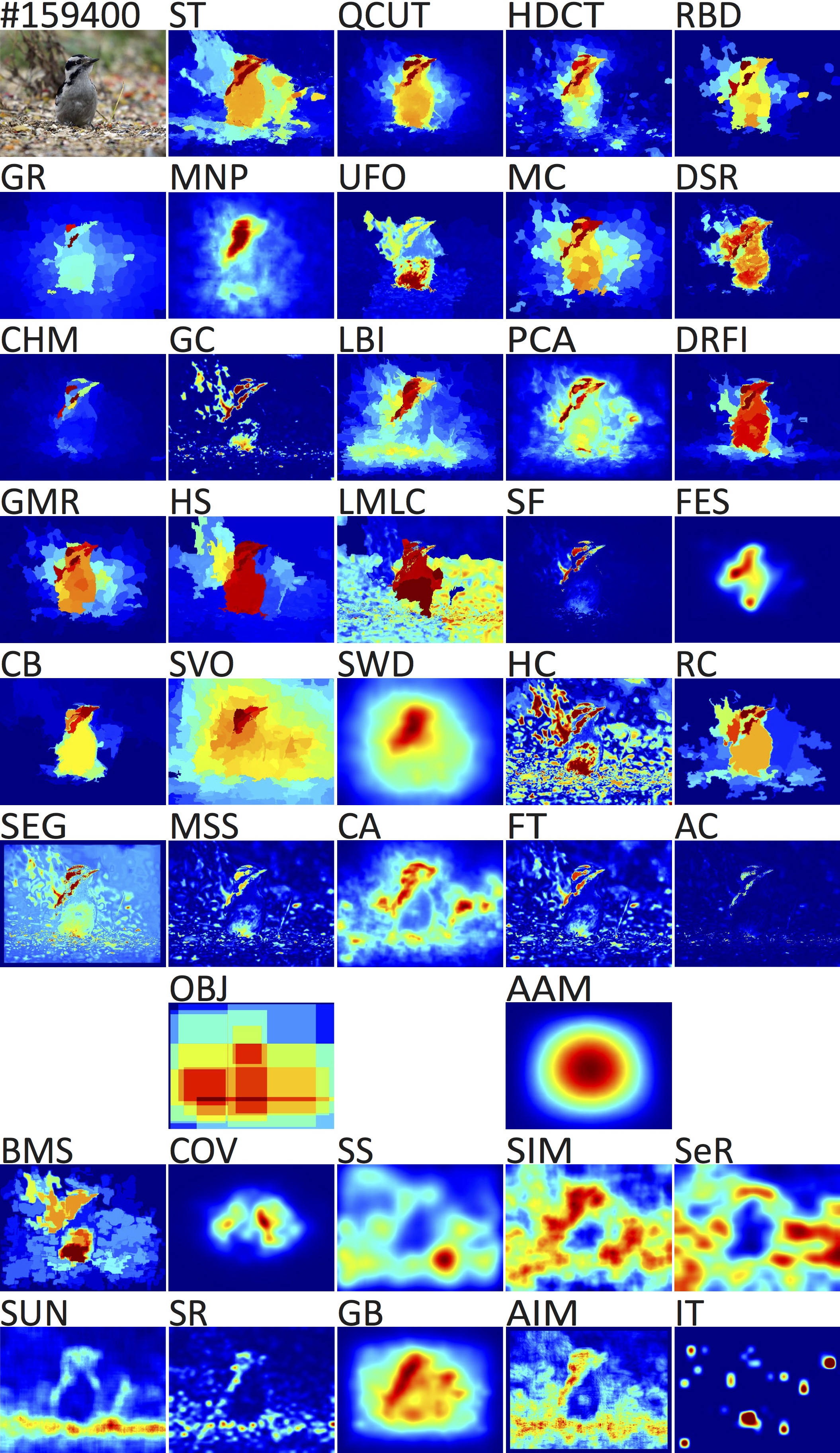} \\
    \caption{Estimated saliency maps from various salient object detection models, object proposal generation model, average annotation map,
        and fixation prediction models.
    }\label{fig:SampleMaps}
\end{figure}

\figref{fig:SampleMaps} shows output maps of all models for a
sample image with relatively complex background.
Dark blue areas are less salient while dark red indicates higher saliency values.
Compared with other models, top contenders like DRFI and DSR suppress most of
the background well while almost successfully detect the whole salient object.
They thus generate higher precision scores and less false positive rates.
Some models that include a center-bias component also result in appealing maps,
\eg, CB.
Interestingly, region-based approaches, \eg, RC, HS, DRFI, BMR, CB, and DSR
always preserve the object boundary well compared with other pixel-based or patch-based models.

We can also clearly see the distinctness of different categories of models.
Salient object detection models try to highlight
the whole salient object and suppress the background.
Fixation prediction models often produce blob-like and sparse saliency maps
corresponding to the fixation areas of humans on scenes.
The objectness map is a rough indication of the salient object.
The output of the latter two types of models might not be suitable for segmenting the whole salient object well.

\section{Performance Analysis}
Based on the performances reported above, we also conduct several experiments to provide a detailed analysis of all the benchmarking models and datasets.

\subsection{Analysis of Segmentation Methods}\label{sec:SalCut}

In many computer vision and graphics applications,
segmenting regions of interest is of great practical importance
\cite{li2013partial,chen2009sketch2photo,chia2011semantic,zhu2012unsupervised,he2012mobile,liu2012web,huang2011arcimboldo}.
The simplest way of segmenting a salient object is to binarize
the saliency map using a fixed threshold, which might be hard to choose.
In this section, we extensively evaluate two additional most commonly
used salient object segmentation methods,
including adaptive threshold \cite{achanta2009frequency} and SaliencyCut \cite{ChengPAMI}.
Average $F_\beta$ scores for salient object segmentation results on
seven benchmark datasets are shown in~\figref{fig:FMeasure}.
Each segmentation algorithm was fed with saliency maps produced
by all 41 compared models.

Except \textbf{JuddDB}, \textbf{PASCAL-S} and \textbf{SED2} datasets,
best segmentation results are all achieved via
SaliencyCut method combined with a sophisticated
salient object detection model (\eg, DRFI, RBD, MNP).
This suggests that enforcing label consistency in terms of
using graph-based segmentation and global appearance statistics
benefits salient object segmentations.
The default SaliencyCut \cite{ChengPAMI} program only outputs
the most dominate salient object,
This causes results for \textbf{SED2}, \textbf{PASCAL-S} and \textbf{JuddDB}
benchmarks to be less optimal,
as images in these two datasets (see \figref{fig:SampleImgs})
do not follow the ``single none ambiguous salient object assumption''
made in \cite{ChengPAMI}.

As also observed by most works in image segmentation literature,
nearby pixels with similar appearance tend to have similar object labels.
To validate this, we demonstrated in \figref{fig:SampleCut}(a) some better segmentation
results by further enforcing
label consistency among nearby and similar pixels.
Enforcing such label consistency often helps improve
labeling pixels specially when the majority of the salient object pixels have been highlighted in the
detection phase.
Challenging examples might still exist, however,
such as complex object topology, spindle components,
and similar appearance with respect to image background.
More results of using the best combination,
DRFI saliency maps and SaliencyCut segmentation,
are demonstrated for images with various complexities,
as shown in \figref{fig:SampleCut}(b).

\newcommand{\AddImgW}[1]{\includegraphics[width=0.189\linewidth]{SampleResults/#1}}
\renewcommand{\AddImgs}[1]{\AddImgW{#1.jpg} \AddImgW{#1_DRFI} \AddImgW{#1_DRFI_FT} \AddImgW{#1_DRFI_SC} \AddImgW{#1.png}}
\renewcommand{\AddImg}[2]{\includegraphics[height=#1\linewidth]{SampleResults/#2}}
\newcommand{\addCutResult}[2]{\AddImg{#1}{#2} \AddImg{#1}{#2_DRFI_SC}}
\begin{figure}
  \centering
  \AddImgs{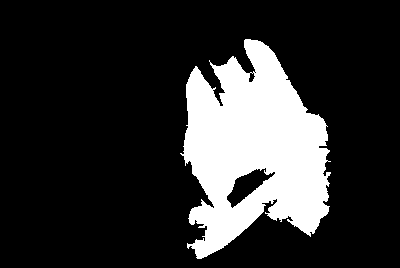} \\ \vspace{0.03in}
  \AddImgs{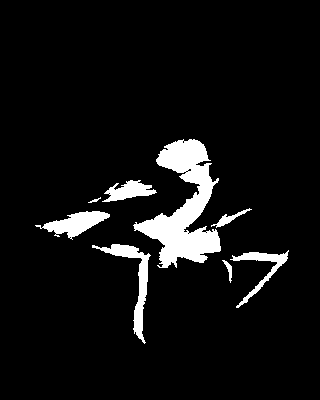} \\ \vspace{0.03in}
  \AddImgs{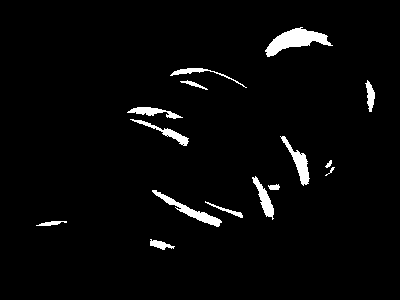} \\
  {\small (a) Left to right: image, saliency map, AdpT: Adaptive Threshold, SCut: SaliencyCut and gTruth: Ground Truth.} \\ \vspace{.03in}
  \addCutResult{0.119}{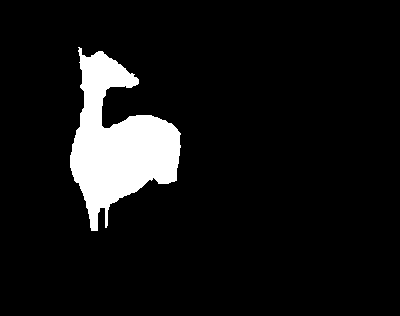}
  \addCutResult{0.119}{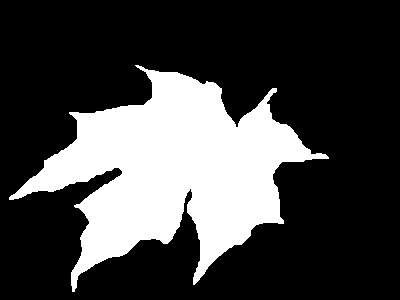}
  \addCutResult{0.119}{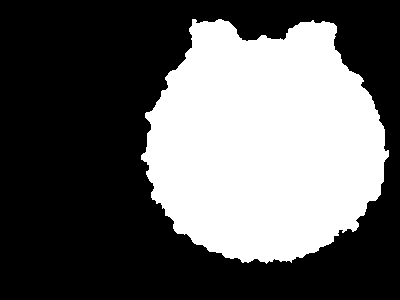} \\ \vspace{0.03in}
  \addCutResult{0.113}{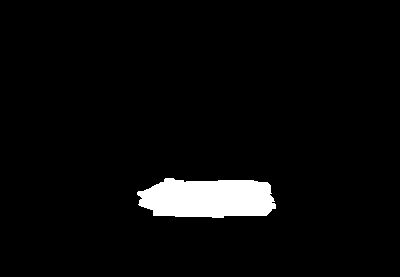}
  \addCutResult{0.113}{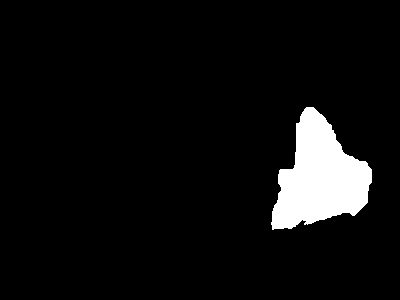}
  \addCutResult{0.113}{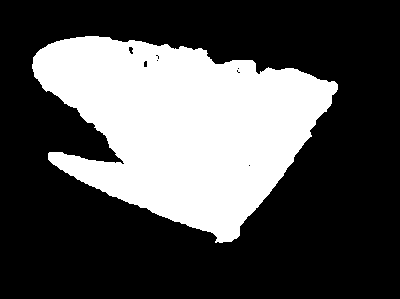} \\ \vspace{0.03in}
  \addCutResult{0.117}{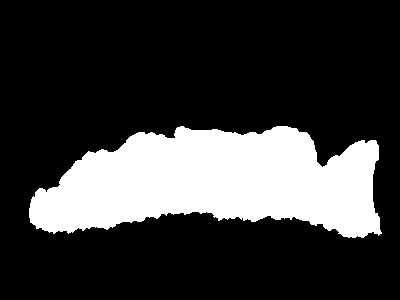}
  \addCutResult{0.117}{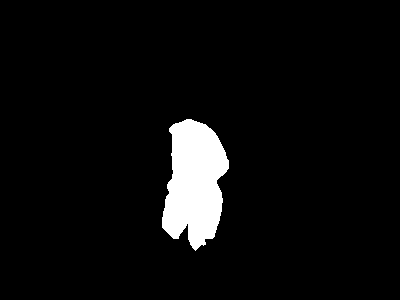}
  \addCutResult{0.117}{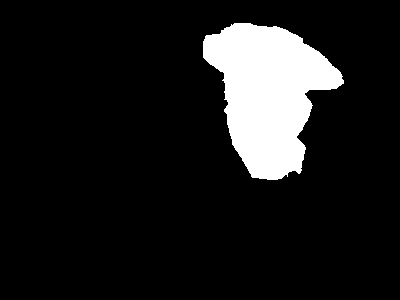}\\
  {\small (b) DRFI model output fed to the SaliencyCut algorithm.}
  \caption{Samples of salient object segmentation results.
  }\label{fig:SampleCut}
\end{figure}

A failure case of SaliencyCut segmentation along with intermediate results
is also shown in the last row of~\figref{fig:SampleCut}(a).
Due to the complex topology of the salient object,
label consistency in a local range considered in the SaliencyCut algorithm may
not work well.
Additionally, the appearance of the object looks very distinct due to the existence
of shading and reflection,
which makes the segmentation of the whole object very challenging.
Therefore, only a part of the object is finally segmented.

\subsection{Analysis of Center Bias}

\begin{figure}[t]
    \centering	
    \includegraphics[width=\linewidth]{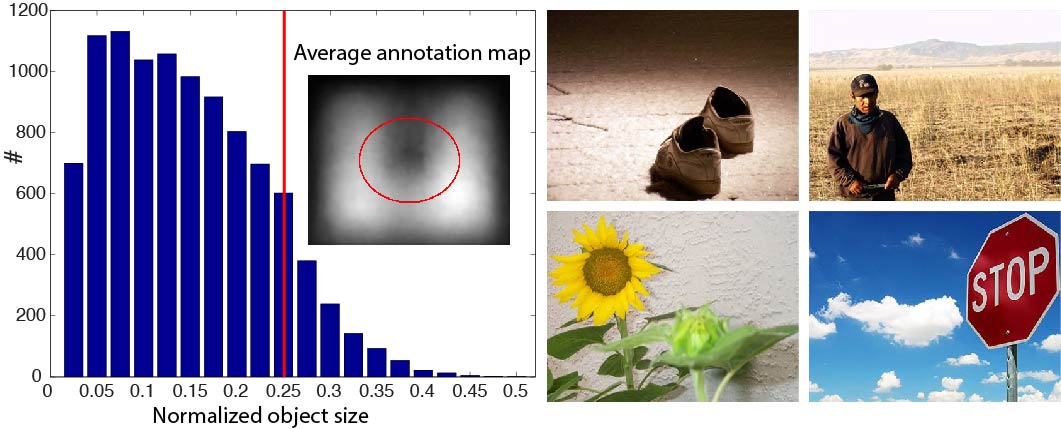} \\
    \caption{Left: Histogram of object center over all images, threshold (red line = 0.247),
        and annotation map over 1000 less center-biased images from \textbf{MSRA10K} dataset.
        Right: Four less center-biased images. The overlaid circle illustrates the center-bias threshold.
    }\label{fig:CBSamples}
\end{figure}

In this section, we study the center-bias challenge since it has caused a major problem
in evaluating fixation prediction and salient object detection models.
Some studies usually add a Gaussian center prior to models when comparing them. This might not be fair as several salient object detection models already contain
center-bias at different levels. Alternatively, we randomly choose 1000 images with no/less center bias
from the \textbf{MSRA10K} dataset.
First, the distance of salient object centroid to the image center is computed
for each image.
Those images for which such distance is bigger than a threshold are then chosen.
Some sample images with no/less center-bias,
as well as an illustration of the threshold of choosing images,
are shown in \figref{fig:CBSamples}.
The average annotation of less center-biased images shows two
peaks on the left and on the right of the image,
which is suitable for testing the performance of salient object detection models
on off-center images.

\newcommand{\addFigA}[3]{\begin{overpic}[width=.49\linewidth,height=6.5cm]{ResNew/UnCentered_#2.pdf}\put(#3,9){#1}\end{overpic}}
\begin{figure*}[!htbp]
    \hspace{-.05in}
    \addFigA{}{Roc}{33} \hspace{.045in}
    \addFigA{}{Pr}{36} \hspace{.05in} \\
    \renewcommand{\arraystretch}{1.2}
    \small
    \renewcommand{\tabcolsep}{0.688mm}
    \begin{tabular}{|l||c|c|c|c|c|c|c|c|c|c|c|c|c|c|c|c|c|c|c|c|} \hline
	Method &   ST& QCUT& HDCT&  RBD&   GR&  MNP&  UFO&   MC&  DSR&  CHM&   GC&  LBI&  PCA& DRFI&  GMR&   HS& LMLC&   SF&  FES&   CB\\\hline
	Max   & .798 & .808 & \second{.822} & .811 & .791 & .661 & .805 & .764 & .776 & .746 & .697 & .685 & .750 & \first{.831} & .754 & \third{.815} & .720 & .747 & .621 & .693 \\
	AUC   & .941 & .911 & .941 & \second{.943} & .925 & .912 & .929 & .888 & .938 & .920 & .860 & .910 & .928 & \first{.964} & .886 & .918 & .896 & .885 & .839 & .872 \\
	MAE   & .135 & \second{.115} & .122 & \first{.106} & .183 & .188 & .128 & .171 & \third{.117} & .138 & .164 & .197 & .162 & .127 & .148 & .150 & .201 & .150 & .160 & .207 \\
\hline
\end{tabular}
\renewcommand{\tabcolsep}{0.75mm}
\begin{tabular}{|l||c|c|c|c|c|c|c|c|c|c|c|c|c|c|c|c|c|c|c|c|c|} \hline
	Method &  SVO&  SWD&   HC&   RC&  SEG&  MSS&   CA&   FT&   AC&  OBJ&  BMS&  COV&   SS&  SIM&  SeR&  SUN&   SR&   GB&  AIM&   IT & AAM\\\hline
	Max   & .792 & .521 & .700 & .744 & .629 & .666 & .620 & .671 & .521 & .708 & .739 & .463 & .571 & .515 & .546 & .498 & .444 & .590 & .540  & .460 & .328\\
	AUC   & \third{.942} & .813 & .898 & .855 & .828 & .868 & .896 & .843 & .800 & .915 & .879 & .805 & .852 & .858 & .849 & .795 & .750 & .850 & .836 & .655 & .716 \\
	MAE   & .325 & .291 & .176 & .177 & .300 & .167 & .199 & .183 & .177 & .243 & .146 & .176 & .225 & .363 & .273 & .276 & .184 & .208 & .265  & .165 & .406 \\
\hline
\end{tabular}
    \caption{Results of center-bias analysis over 1000 less center-biased images
        chosen from the \textbf{MSRA10K} dataset.
        Top: ROC and PR curves, Bottom: Max $F_\beta$, AUC, and MAE scores for all models.
    }\label{fig:CenterBias}
\end{figure*}

We evaluate all the compared 41 models on these 1000 images.
PR and ROC curves, $F_\beta$, AUC, and MAE scores are all shown in \figref{fig:CenterBias}.
DRFI and DSR again perform the best.
Overall, most models' performance decrease when testing on no/less center biased images
(\eg, the AUC score of MC declines from 0.951 to 0.888),
while a few others show increase.
For example, the AUC score of SVO raises from 0.930 to 0.942 and it gets the second ranking.
Some models, \eg, HS (with the second ranking in terms of $F_\beta$ score),
performs better according to their rank changes w.r.t the whole \textbf{MSRA10K} dataset.
DRFI still wins over other models here with a large margin.
The difference in $F_\beta$, AUC, and MAE scores are not very large for this model over all data
and 1000 less center-biased images (difference are $0.05$, $0.05$, and $0.009$, respectively).
This means that this model is not taking advantage of center-bias much.
In the contrast, CB model uses a great deal of location prior and that is why it's performance drops heavily
when applied to the off-center images (difference are $0.122$, $0.122$, and $0.029$, respectively).

Additionally, it can be observed from~\figref{fig:AverageMap}(f),
there is less center bias over the \textbf{SED2} dataset where
there is less activation in the center of its average annotation map.
We can therefore study the center bias on it.
Similarly, DRFI and DSR outperform other models in terms of $F_\beta$,
AUC, and MAE scores,
indicating they are more robust to the location variations of salient objects.
HS again ranks second according to the $F_\beta$ score.
Fig.~\ref{fig:easyDiffUncentered} shows best and worst off-centered stimuli for DRFI and DSR models.

Overall, all the models perform well above the chance level over
either the less center-biased subset of \textbf{MSRA10K} or \textbf{SED2}.
It is also worth noticing that the AAM model performs significantly
worse on these two datasets, as well as \textbf{JuddDB},
validating our motivation of studying center bias on them.


\graphicspath{{./BestWorst/}}
\renewcommand{\AddImg}[1]{\includegraphics[width=0.95\linewidth]{#1}}
\renewcommand{\AddImgsU}[2]{\begin{sideways} #2 \end{sideways} & \AddImg{#1}}
\begin{figure}[t!]
    \centering
    \begin{tabular*}{\linewidth}{cc}
        \AddImgsU{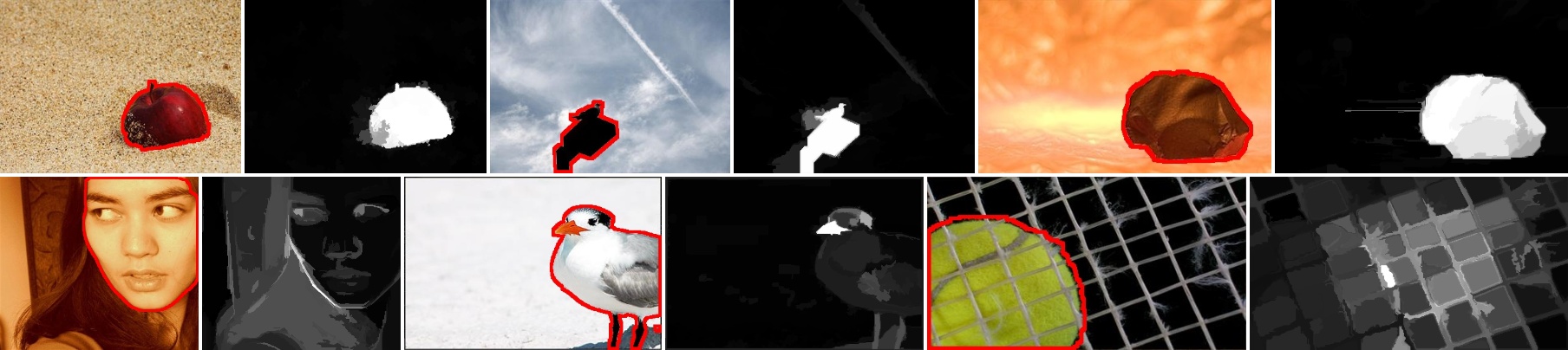}{~DRFI} \\
        \AddImgsU{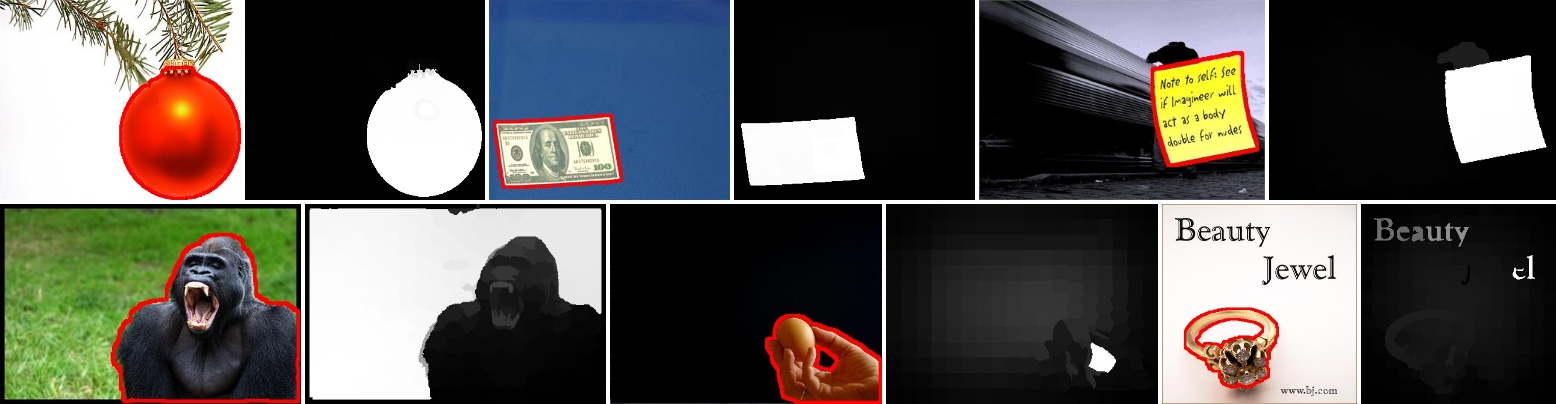}{~DSR} \\
    \end{tabular*}
    \caption{Top and Bottom rows for each model illustrate best and worst cases in off-centered images.
    }\label{fig:easyDiffUncentered}
\end{figure}




\newcommand{\AddImgWW}[1]{\includegraphics[width=0.189\linewidth]{Background/#1}}
\renewcommand{\AddImgs}[1]{\AddImgWW{#1.jpg} \AddImgWW{#1_DRFI} \AddImgWW{#1_DSR} \AddImgWW{#1_MC} \AddImgWW{#1_IT}}
\begin{figure}[t!]
  \centering
  \AddImgs{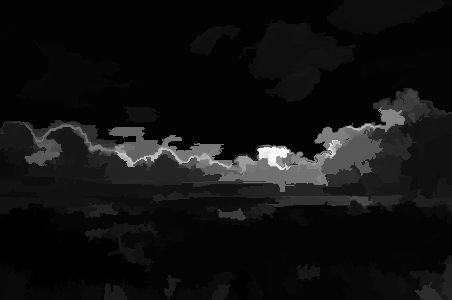} \\ \vspace{0.03in}
  \AddImgs{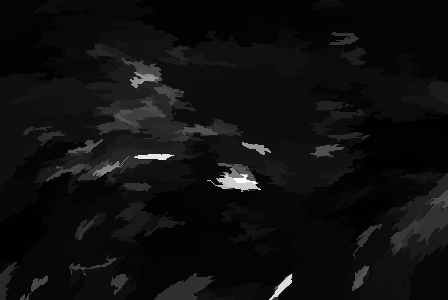} \\ \vspace{0.03in}
  \AddImgs{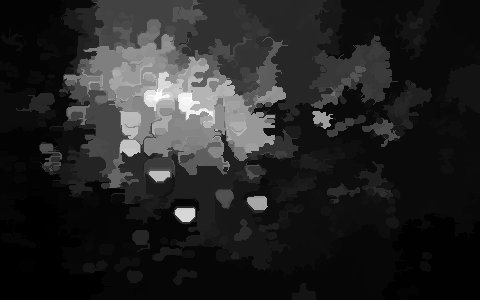} \\ \vspace{0.03in}
    \AddImgs{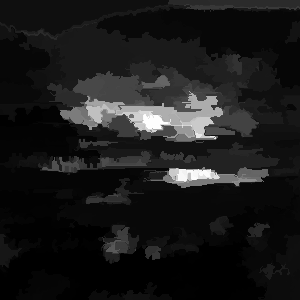} \\ \vspace{0.03in}
      \AddImgs{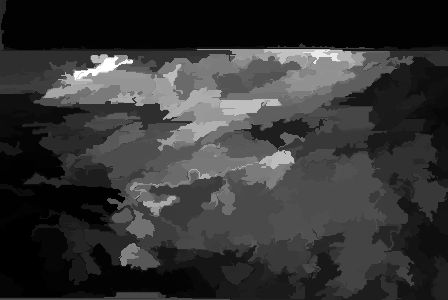} \\ \vspace{0.03in}
      \AddImgs{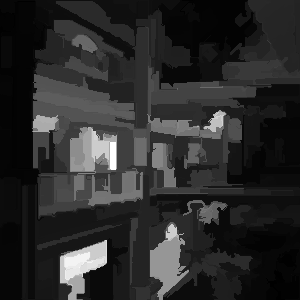} \\
  \caption{Sample background-only images and prediction maps of DRFI, DSR, MC, and IT models.
  }\label{fig:SampleBG}
\end{figure}

\subsection{Analysis of Salient Object Existence}
The existence of a salient object in the image is somewhat neglected by the community.
Almost all of existing salient object detection models assume
that there is at least one salient object in the input image.
This impractical assumption might lead to less optimal performance on ``background images'',
which do not contain any dominant salient objects,
as studied in~\cite{wang2012salient}. Just recently, Zhang~\etal~\cite{zhang2015salient} introduced a fast method for a more challenging task of counting (subitizing) salient objects in a scene.

We can see from~\figref{fig:SampleBG} that no dominated salient object exists in background images consisting of only textures or cluttered backgrounds.
A good model should generate a dark (blank) saliency map on a background image, \ie,
without any activation as there are no salient objects.
\figref{fig:SampleBG} shows
saliency maps using three top salient object detection models
and a classical fixation prediction model on background images.
Top salient object detection models like DRFI, DSR, and MC do not perform well
and often generate activations on the background images even though only regular
textures exist (the second and third rows of~\figref{fig:SampleBG}).
This is reasonable as they always assume there exist salient objects in
the input image and will try their best to find one.
These models can be distracted by the clutter in the background
since high contrast always exist on the cluttered region.
Most of existing salient object detection models compute saliency based on contrast values.
These cluttered regions are thus more likely considered as salient.
It is worth pointing out that ground truth of \emph{eye fixations} do exist on such background images.

In addition to salient object existence, quantitative evaluations of models on background images is an open problem as well.
Note that it is not feasible to calculate PR and ROC curves (and thus $F_{\beta}$ and AUC scores)
on background images since the ground truth positive labeling is empty.
MAE score is not informative either as most salient object detection methods explicitly normalize
the saliency maps in the range of [0,255] as a post-processing step.
By demonstrating qualitative results of salient object detection models on some background images,
we aim to motivate future works focusing on salient object detection on background images.



\subsection{Analysis of Worst and Best Cases for Top Models}
To understand what are the challenges for existing salient object detection models,
we illustrate the three best and the three worst cases for top models over all seven benchmark datasets.
The stimuli for 11 top models were sorted according to the $F_\beta$ scores.
We only give a demonstration of DRFI and MC models in~\figref{fig:easyDiff} due to limited space.
See our online challenge website for additional illustrations.


It can be noticed from~\figref{fig:easyDiff} that models share the same easy
and difficult stimuli.
Both DRFI and MC perform substantially well on the cases
where a dominated salient object exists in a relatively clean background.
Since most existing salient object detection models do not utilize any
high-level prior knowledge,
they may fail when a complex scene has a cluttered background or
when the salient object is \emph{semantically} salient
(\eg, DRFI fails on images with faces in \textbf{MSRA10K}).
Another reason causing poor saliency detection is object size.
Both DRFI and MC models have difficulty in detecting small objects (See hard cases on \textbf{DUT-OMRON} and \textbf{JuddDB}).

Particularly, since saliency cues adopted by DRFI are mainly based on contrast,
this model fails on scenes where salient objects share close appearance with the
background (\eg, the hard cases of \textbf{MSRA10K} and \textbf{ECSSD}).
Another possible reason is related to the failure in segmenting the image.
MC relies on the pseudo-background prior that the image border areas are background.
That is why it fails on scenes where the salient object touches the image border,
\eg, the gorilla image in \textbf{MSRA10K} dataset (4th row of the right column of~\figref{fig:easyDiff}).

\graphicspath{{./BestWorst/}}
\renewcommand{\AddImg}[1]{\includegraphics[width=0.48\linewidth,height=0.129\linewidth]{#1}}
\renewcommand{\AddImgs}[2]{\begin{sideways} #2 \end{sideways} & \AddImg{#1_DRFI} && \AddImg{#1_MC}}
\begin{figure*}[!htbp]
    \centering
    \begin{tabular*}{\linewidth}{cccc}
        \AddImgs{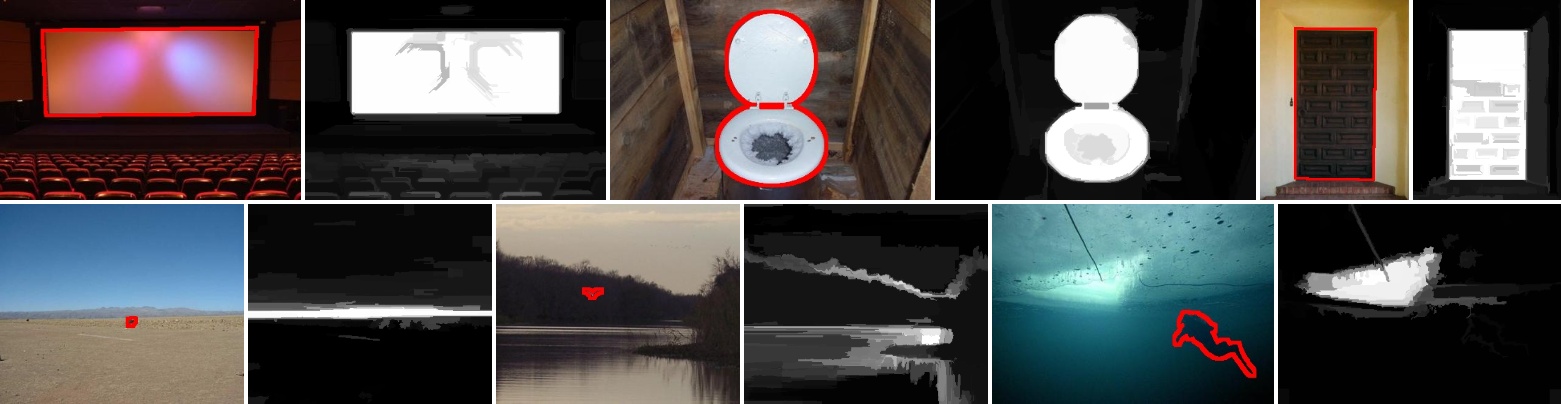}{~\textbf{DUT-OMRON}} \\
        \AddImgs{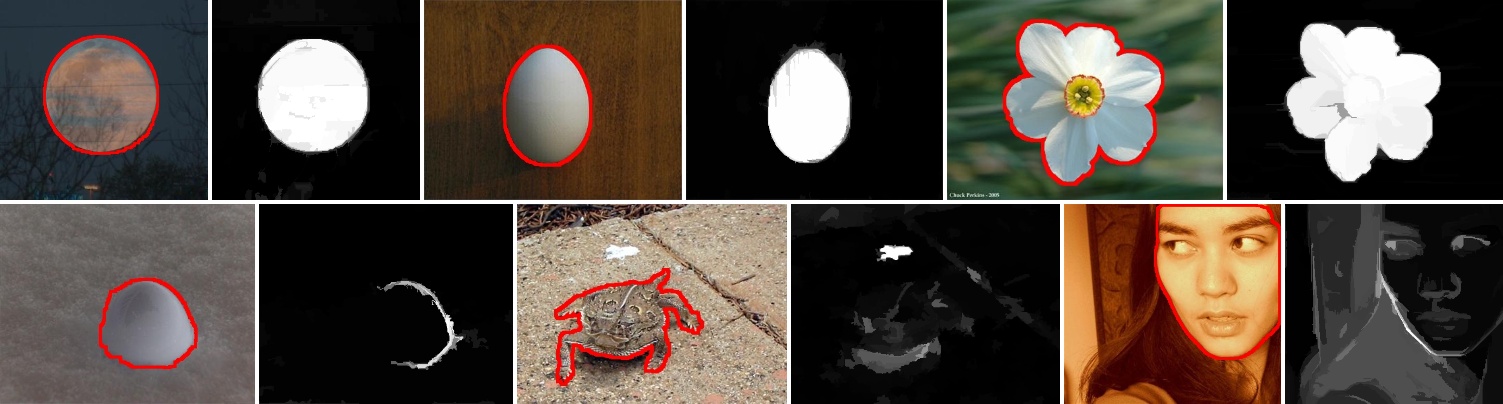}{~ ~ \textbf{MSRA10K}} \\
        \AddImgs{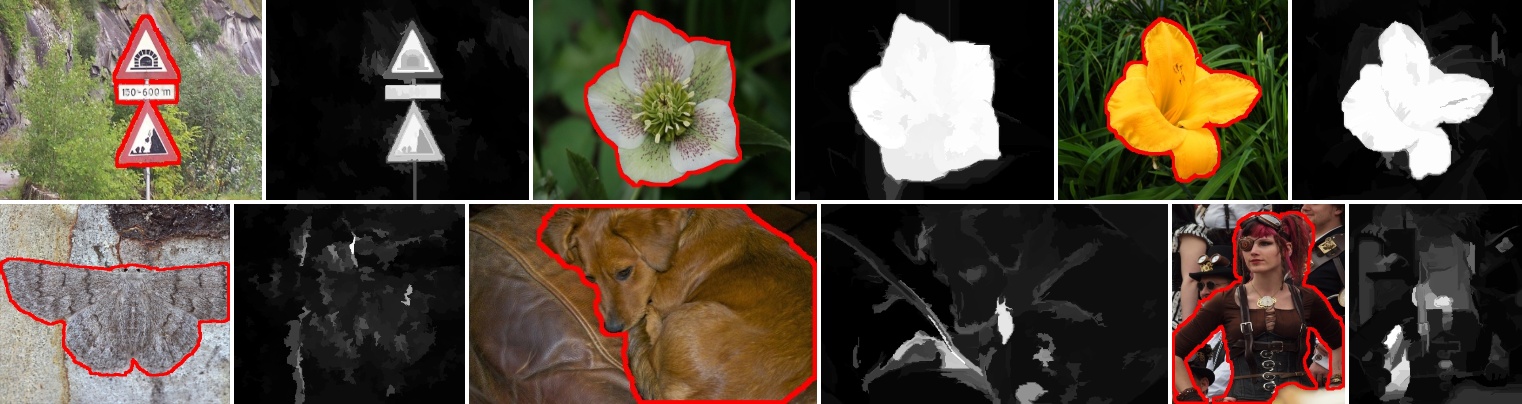}{~ ~ ~ \textbf{ECSSD}} \\
        \AddImgs{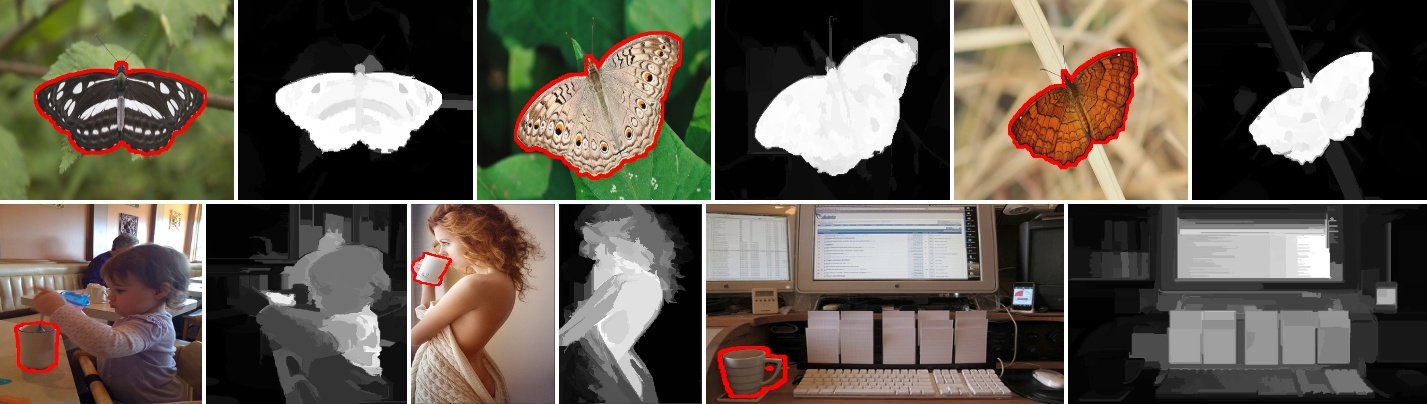}{~ ~ \textbf{THUR15K}} \\
        \AddImgs{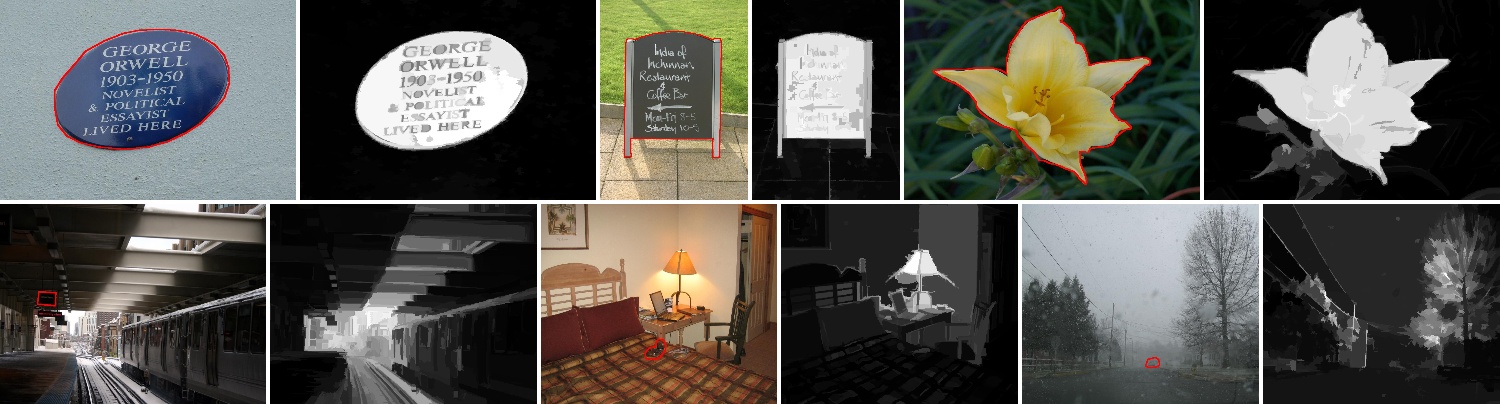}{~ ~ \textbf{JuddDB}} \\
        \AddImgs{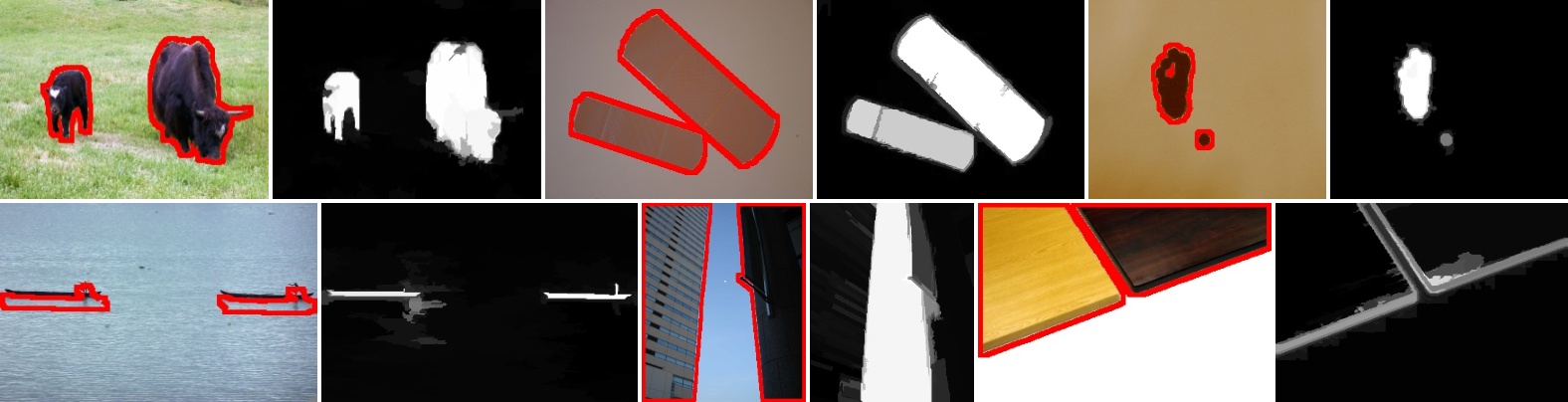}{~ ~ ~ ~ \textbf{SED2}} \\
        \AddImgs{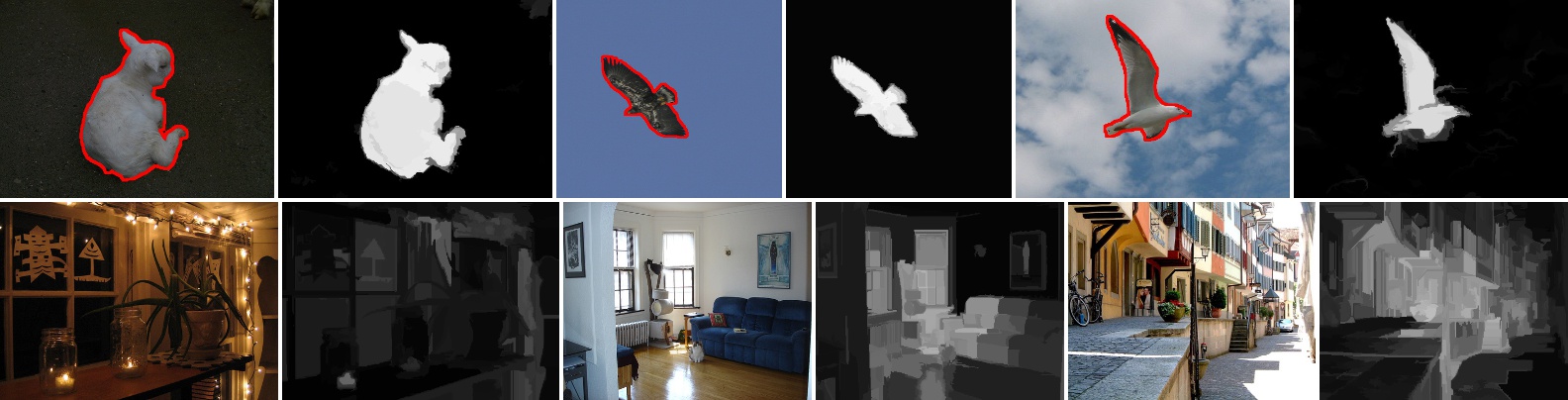}{~ ~ \textbf{PASCAL-S}} \\
        & (a) DRFI  & ~ &  (b) MC
    \end{tabular*}
    \caption{Best (1st rows for each model on a dataset) and worst (2nd rows)
        cases of DRFI and MC.
        Ground-truth object(s) is denoted by a red contour.
    }\label{fig:easyDiff}
\end{figure*}

\begin{figure}
\centering
\includegraphics[width=\linewidth]{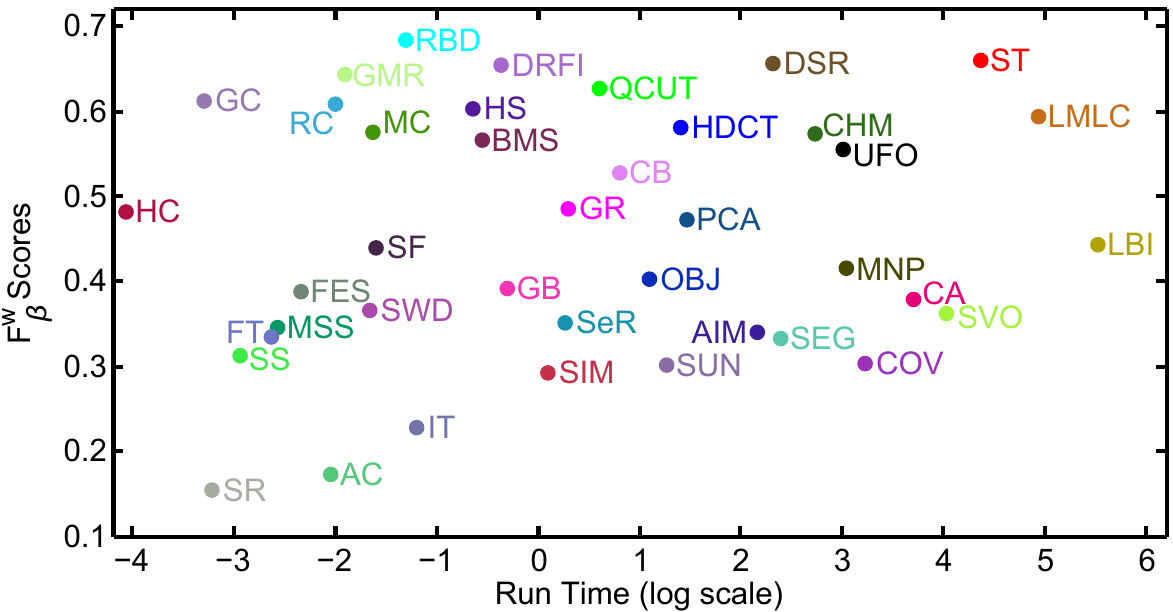}
\caption{$F_\beta^w$ scores versus (log scale of) runtime of different methods based on the quantitative results of MSRA10K dataset.}
\label{fig:FVsRuntime}
\end{figure}

\subsection{Runtime Analysis}

Runtime of compared models are shown in
\tabref{tab:salientObjModelsIntrin} over all 10K images
of \textbf{MSRA10K} (typical image resolution of $400\times 300$)
using an Intel Xeon E5645 2.40GHz CPU with 8 GB RAM.
A 2D scatter plot of $F_\beta^w$ scores versus running time of different methods based on the quantitative results of MSRA10K dataset is shown in~\figref{fig:FVsRuntime}, which is helpful to demonstrate the trade-off between efficacy and efficiency of compared models.
%
%

Of all compared methods, the HC model is the fastest (about 0.017 seconds per image)
followed by GC and SR models.
The best model in our benchmark (DRFI) needs about 0.697
seconds to process one image.
We can also observe that RC, GMR, MC, and RBD share similar trade-offs between $F_\beta^w$ scores and runtime.

\section{Discussions and Conclusions}

From the results obtained so far, we summarize in \tabref{tab:summaryRank} the rankings of models based on average performance over datasets in terms of segmentation methods, center bias, salient object existence, and run time\footnote{We have created a unified repository for sharing code and data where researchers can run models with a single
click or can add new models for benchmarking purposes. All codes, data, and results are available in our online benchmark website:
\href{http://mmcheng.net/salobjbenchmark/}{http://mmcheng.net/salobjbenchmark/}}. Based on the rankings, we conclude that:

``\emph{DRFI, QCUT, RBD, ST, DSR, and MC are the top 6 models for salient object detection.}''

\begin{figure}
\includegraphics[width=\linewidth]{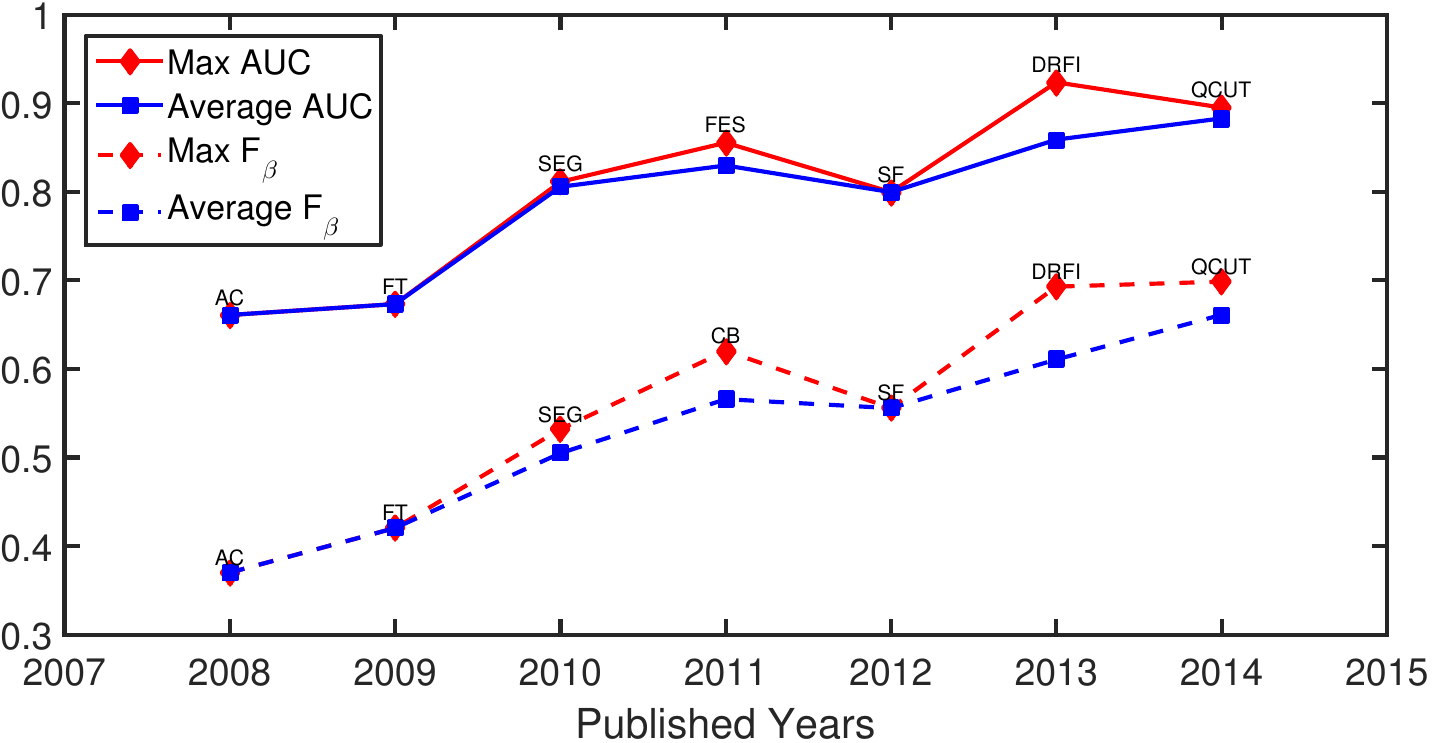}
\caption{Maximum and average AUC and $F_{\beta}$ scores of different salient object methods versus their publication years. Model accuracy shows an increasing trend.}
\label{fig:performanceTrend}
\end{figure}

To gauge the progress in this field, we show in \figref{fig:performanceTrend}, the maximum and average AUC and $F_{\beta}$ scores of different salient object detection methods versus their publication years. We find a continuous ascending success rate over the last couple of years which raises the hope that even better salient object detection models are possible in the future.

\begin{figure*}[!htbp]
    \renewcommand{\arraystretch}{1}
    \renewcommand{\tabcolsep}{1.272mm}
    \footnotesize
    \begin{tabular}{|l||c|c|c|c|c|c|c|c|c|c|c|c|c|c|c|c|c|c|c|c|} \hline
	Method &   ST& QCUT& HDCT&  RBD&   GR&  MNP&  UFO&   MC&  DSR&  CHM&   GC&  LBI&  PCA& DRFI&  GMR&   HS& LMLC&   SF&  FES&   CB\\\hline
$F_\beta$& \third{3} & \second{2} & 9 & 5 & 14 & 29 & 13 & 4 & 6 & 10 & 20 & 26 & 16 & \first{1} & 7 & 8 & 19 & 21 & 18 & 12 \\
$F_\beta^w$& \second{2} & 6 & 11 & \first{1} & 17 & 20 & 15 & 10 & 4 & 12 & 8 & 18 & 19 & \third{3} & 5 & 7 & 13 & 25 & 26 & 16 \\
	AUC   & \third{3} & 4 & 9 & 5 & 23 & 22 & 20 & 6 & \second{2} & 8 & 31 & 15 & 7 & \first{1} & 11 & 12 & 26 & 28 & 19 & 21 \\
	MAE   & 7 & \first{1} & 6 & \third{3} & 27 & 29 & 9 & 12 & \second{2} & 5 & 17 & 25 & 18 & 4 & 10 & 19 & 26 & 15 & 8 & 23 \\
	AdpT  & 7 & \first{1} & 8 & 6 & 19 & 23 & 15 & 4 & \third{3} & 9 & 24 & 20 & 14 & \second{2} & 5 & 16 & 27 & 21 & 12 & 18 \\
	SCut  & \first{1} & 8 & 6 & \third{3} & 26 & 18 & 12 & 15 & 21 & 10 & 31 & 9 & 16 & \second{2} & 17 & 7 & 19 & 36 & 34 & 11 \\
	CB    & 7 & 5 & \second{2} & 4 & 9 & 25 & 6 & 11 & 10 & 15 & 20 & 22 & 13 & \first{1} & 12 & \third{3} & 18 & 14 & 27 & 21 \\
	Time  & 38 & 23 & 27 & 14 & 22 & 34 & 33 & 12 & 30 & 32 & \second{2} & 40 & 28 & 18 & 10 & 16 & 39 & 13 & 7 & 24 \\
	\hline
	Overall Rank & 4 & \second{2} & 9 & \third{3} & 18 & 23 & 12 & 6 & 5 & 8 & 19 & 20 & 15 & \first{1} & 7 & 10 & 17 & 29 & 22 & 14 \\
	\hline
\hline
\end{tabular}
    \renewcommand{\tabcolsep}{1.379mm}
\begin{tabular}{|l||c|c|c|c|c|c|c|c|c|c|c|c|c|c|c|c|c|c|c|c|c|} \hline
	Method &  SVO&  SWD&   HC&   RC&  SEG&  MSS&   CA&   FT&   AC&  OBJ&  BMS&  COV&   SS&  SIM&  SeR&  SUN&   SR&   GB&  AIM&   IT & AAM\\\hline
$F_\beta$& 17 & 22 & 32 & 11 & 25 & 28 & 31 & 34 & 37 & 27 & 15 & 24 & 33 & 39 & 36 & 40 & 41 & 23 & 35 & 38 & 30\\
$F_\beta^w$ & 30 & 27 & 24 & 9  & 31 & 34 & 28 & 37 & 39 & 21 & 14 & 33 & 36 & 35 & 29 & 38 & 40 & 22 & 32 & 41 & 23\\
	AUC   & 13 & 18 & 33 & 17 & 30 & 29 & 27 & 40 & 38 & 24 & 16 & 10 & 35 & 34 & 36 & 39 & 37 & 14 & 32 & 41 & 25\\
	MAE   & 40 & 34 & 31 & 14 & 37 & 16 & 30 & 28 & 22 & 36 & 13 & 11 & 32 & 41 & 39 & 38 & 20 & 24 & 35 & 21 & 33\\
	AdpT  & 38 & 22 & 32 & 13 & 39 & 26 & 28 & 34 & 40 & 29 & 10 & 11 & 30 & 35 & 33 & 41 & 36 & 17 & 37 & 31 & 25\\
	SCut  & 13 & 4  & 32 & 14 & 24 & 38 & 29 & 37 & 39 & 20 & 23 & 35 & 30 & 25 & 27 & 33 & 40 & 5  & 28 & 41 & 22\\
	CB    & 8  & 35 & 19 & 16 & 26 & 24 & 28 & 23 & 34 & 31 & 17 & 38 & 30 & 36 & 32 & 37 & 40 & 29 & 33 & 39 & 41\\
	Time  & 37 & 11 & \first{1} & 9 & 31 & 6 & 36 & 5 & 8 & 25 & 17 & 35 & 4 & 20 & 21 & 26 & \third{3} & 19 & 29 & 15 & -\\
	\hline
	Overall Rank & 27 & 21 & 35 & 13 & 30 & 33 & 28 & 38 & 41 & 26 & 11 & 25 & 31 & 36 & 34 & 37 & 39 & 16 & 32 & 40 & 24 \\
\hline
\end{tabular} 
    \caption{Summary rankings of models under different evaluation metrics over all datasets (excluding SED2).
The overall rankings of different methods are computed based on the average (the higher the better) of AUC, (1-MAE), Max $F_{\beta}$, AdpT, ScutT, and $F_{\beta}^w$ scores.
The top three models under each evaluation metric are highlighted in red, green and blue.
    }\label{tab:summaryRank}
\end{figure*}

By investigating the performances and the design choices of all compared models, our extensive evaluations do suggest some clear messages about commonly used design choices, which could be valuable for developing future algorithms. We refer readers to our recent survey~\cite{borji2014salient} for a comprehensive review of different design choices adopted for salient object detection.

\begin{itemize}\itemsep5pt


\item From the elements perspective, top five models (except QCUT) are built upon superpixels (regions). On the one hand, compared with pixels, more effective features (\eg, color histogram) can be extracted from regions. On the other hand, compared with patches, the boundary of the salient object is better preserved for region-based approaches, leading to more accurate detection performance. Moreover, since the number of superpixels is far less than the number of pixels or patches, region-based methods has the potential to run faster.

\item All the top six models explicitly consider the background prior, which assumes that the area in the narrow border of the image belongs to the background. Compared with the location prior of a salient object, such a background prior performs more robust.

\item The leading method in our benchmark (\ie,~DRFI),
discriminatively trains a regression model to predict region saliency according to
a 93-dimensional feature vector. Instead of purely relying on the cues extracted only from the input image, DRFI resorts to human annotations to automatically discover feature integration rules. The high performance of this simple learning-based method encourages pursuing data-driven approaches for salient object detection.

\end{itemize}




However, even considering top performing models, salient object detection still seems far from being solved. To achieve more appealing results, three challenges should be addressed.
First, in our large-scale benchmark (see \secref{sec:benchmark}),
all top performing algorithms use the location prior cues, limiting their adaptation to general cases.
Second, although the ranking of top scoring models are quite consistent across datasets, performance scores ($F_\beta$ and AUC) drop significantly from easier datasets to more difficult ones. The third challenge regards the run time of models. Some models need around one minute to process a $400\times300$ image
(\eg, CA: 40.9s, SVO: 56.5s, and LMLC 140s).



One area for future research would be designing scores for tackling dataset
biases and evaluation of saliency segmentation maps
with respect to ground-truth annotations similar to~\cite{margolinevaluate}. In this benchmark, we only focused on single-input scenarios. Although some RGBD datasets exist~\cite{Peng14rgbd}, benchmark datasets for multiple input images (\eg, salient object detection on videos, co-salient object detection~\cite{borji2014salient}) are still lacking. Another future direction will be following active segmentation algorithms (\eg,~\cite{mishra2012active,borjiTIP2014,liXiaodiCVPR2014}) by segmenting a salient object from a seed point.
For example, a simple model proposed by Borji~\cite{borjiTIP2014} which segments the most salient object (at the peak of a map generated by a fixation prediction model as the seed point) using superpixels outperforms several salient object detection models on scenes with multiple salient objects (JuddDB). This indicates that several models are affected by a bias imposed by some former datasets (i.e., ASD) which is the existence of only one object in the image.
Aggregation of saliency models for building a strong prediction model (similar to~\cite{borji2012salient,MaiNL13Aggregation,le2014saliency}, and behavioral investigation of saliency judgments by humans (\eg,~\cite{borji2013stands,borji2013objects}) are two other interesting directions.
The relationship (similarity and difference) between salient object detection and related fields such as object detection, object proposals, general segmentation, and fixation prediction\footnote{Please see our fixation prediction benchmark at http://saliency.mit.edu} and the ways these areas can benefit from each other still remains to be explored further.

Inspired by the overwhelming performance of deep learning methods in other vision tasks like image classification~\cite{krizhevsky2012imagenet,szegedy2015going} and object detection~\cite{girshick2015rich}, deep convolutional neural networks (CNNs) are also studied in recent works~\cite{zhao2015saliency,SuperCNN_IJCV2015,lin2014saliency,li2015visual}. The leading performance of DRFI demonstrates the effectiveness of data-driven feature integration. Through deep architectures,  more powerful representations can be learned than hand-crafted features for salient object detection tasks even if CNNs are trained for image classification. It indicates the promising direction of investigating deep learning methods for salient object detection in the future.

Saliency models (whether predicting where humans look in a scene or which objects they choose as salient~\cite{borji2013stands,borji2015reconciling}) play an important role in the way we represent and understand scenes at the high level. Saliency models continue to be useful in a variety of domains encompassing human-robot interaction, image processing, and computer vision. So far modeling effort has been focused on improving the performance of existing datasets. State of the art models do very well even on large scale salient object datasets. We believe that it is now the time to consider how saliency detection can help other challenging tasks in computer vision for problems such as describing a scene (e.g., language and vision~\cite{kulkarni2011baby,farhadi2009describing,itti2006attention,antol2015vqa,fang2014captions,zitnick2013learning}), scene understanding (e.g.,~\cite{yun2013studying,antol2015vqa,chen2015microsoft,geman2015visual}), and even object and scene classification (e.g.,~\cite{chen2015microsoft,russakovsky2014imagenet,krizhevsky2012imagenet}).

Salient object detection is a very active research area in computer vision with several papers emerging each year in major conferences and journals. In fact, several models have been introduced since the initial submission of this work. Some, we have included in our benchmark\footnote{We encourage researchers to actively engage in this benchmark and help us gauge the future progress in this field and address potential challenges.} during the review process and some newer ones (such as~\cite{zhao2015saliency,SuperCNN_IJCV2015,lin2014saliency,li2015visual} mainly based on the deep CNNs) will be considered in our online saliency detection benchmark.
We will extensively review and discuss these models in our ongoing work~\cite{borji2014salient}.



\ifCLASSOPTIONcompsoc
  \section*{Acknowledgments}
\else
  \section*{Acknowledgment}
\fi
Authors would like to thank anonymous reviewers for their helpful comments on the paper.
Ali Borji was supported by Defense Advanced Research Projects Agency
(NO. HR0011-10-C-0034), the National Science
Foundation (CRCNS grant number BCS-0827764), the General Motors
Corporation, and the Army Research Office (NO. W911NF-08-1-0360). 
Ming-Ming Cheng is supported by the grants from NSFC (NO. 61572264).
Jia Li is supported by the grants from NSFC (NO. 61370113), and Fundamental Research Funds for the Central Universities.

\bibliographystyle{IEEEtran}
\bibliography{egbib}

\vspace{-.4in}
\begin{biography}[{\includegraphics[width=1in,height=1.25in,clip,keepaspectratio]{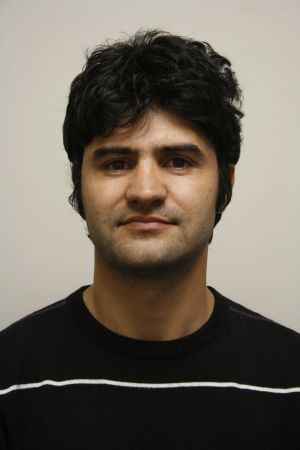}}]{Ali Borji}
received his BS and MS degrees in computer engineering from Petroleum University of Technology, Tehran, Iran, 2001 and Shiraz University, Shiraz, Iran, 2004, respectively. He did his Ph.D. in cognitive neurosciences at Institute for Studies in Fundamental Sciences (IPM) in Tehran, Iran, 2009 and spent four years as a postdoctoral scholar at iLab, University of Southern California from 2010 to 2014. He is currently an assistant professor at University of Wisconsin, Milwaukee. His research interests include visual attention, active learning, object and scene recognition, and cognitive and computational neurosciences.
\end{biography}

\vspace{-.4in}

\begin{biography}[{\includegraphics[width=1in,height=1.25in,clip,keepaspectratio]{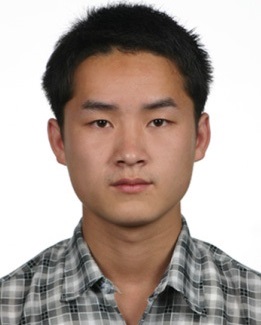}}]
{Ming-Ming Cheng}
received his PhD degree from Tsinghua University in 2012.
Then he did 2 years research fellow, with Prof. Philip Torr in Oxford.
He is now an associate professor at Nankai University.
His research interests includes computer graphics, computer vision, and image processing.
He has received the Google PhD fellowship award, the IBM PhD fellowship award,
and the new PhD Researcher Award from Chinese Ministry of Education.
\end{biography}
\vspace{-.4in}

\begin{biography}[{\includegraphics[width=1in,height=1.25in,clip,keepaspectratio]{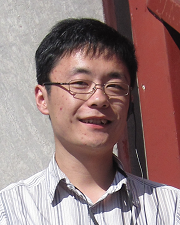}}]{Huaizu Jiang}
is currently working as a research assistant at Institute of Artificial Intelligence
and Robotics, Xi'an Jiaotong University.
Before that, he received his BS and MS degrees from Xi'an Jiaotong University, China,
in 2005 and 2009, respectively.
He is interested in how to teach an intelligent machine to understand the visual scene like a human.
Specifically, his research interests include object detection,
large-scale visual recognition, and (3D) scene understanding.
\end{biography}
\vspace{-.4in}

\begin{biography}[{\includegraphics[width=1in,height=1.25in,clip,keepaspectratio]{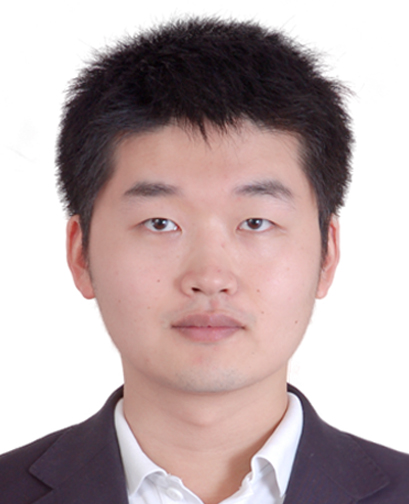}}]{Jia Li}
received his B.E. degree from Tsinghua University in 2005
and Ph.D. degree from the Chinese Academy of Sciences in 2011.
During 2011 and 2013, he was attach to Nanyang Technological University as research fellow and visiting assistant professor, respectively.
He is currently an associate professor at Beihang University, Beijing, China.
His research interests include visual attention/saliency computation, image and video processing, and visual recognition from big data.
\end{biography}

\vfill

\end{document}